\documentclass[11pt, letterpaper, logo, onecolumn, copyright]{main}

\usepackage[authoryear, sort&compress, round]{natbib}

\usepackage[inkscapeformat=png]{svg}

\usepackage[most, breakable, skins]{tcolorbox}

\tcbuselibrary{skins}
\usepackage{lipsum}
\usepackage{tabularx}
\usepackage{afterpage}
\usepackage{booktabs}
\usepackage{subcaption}
\usepackage{makecell}
\usepackage{multirow}
\usepackage{bm}
\usepackage{multicol}
\usepackage{array}
\usepackage{float}
\usepackage{listings, listings-rust}
\usepackage{fontawesome5}
\usepackage{hyperref}
\usepackage{amssymb,graphicx}
\usepackage[dvipsnames]{xcolor}
\usepackage{cleveref}
\usepackage{longtable}
\usepackage{pdflscape}
\usepackage{adjustbox}
\usepackage{nicematrix}
\usepackage{CJKutf8}
\usepackage{ragged2e}
\usepackage{colortbl}
\usepackage{enumitem}
\usepackage[ruled,linesnumbered]{algorithm2e}
\usepackage{pifont}
\usepackage[htt]{hyphenat}
\usepackage{amsmath}
\usepackage{amsthm}  
\usepackage{mathrsfs} 
\usepackage{circledsteps}
\usepackage{diagbox}
\usepackage{bookmark}

\usepackage{arydshln}
\usepackage{bm}

\usepackage{wrapfig}
\usepackage{xspace}
\usepackage{tikz}
\usepackage[normalem]{ulem}

\usepackage{pgfplots}
\usepackage{pgfplotstable}
\pgfplotsset{compat=1.18}

\definecolor{medgray55}{gray}{0.55}
\definecolor{medgray}{gray}{0.7}
\definecolor{litegray}{gray}{0.9}
\definecolor{gblue}{RGB}{210, 227, 252}
\definecolor{gred}{RGB}{250, 210, 207}
\definecolor{gyellow}{RGB}{254, 239, 195}
\definecolor{ggreen}{RGB}{206, 234, 214}
\definecolor{gorange}{RGB}{254, 223, 200}

\definecolor{gblue9}{RGB}{23, 78, 166}
\definecolor{gred9}{RGB}{165, 14, 14}
\definecolor{gyellow9}{RGB}{227, 116, 0}
\definecolor{ggreen9}{RGB}{13, 101, 45}
\definecolor{gorange9}{RGB}{176, 96, 0}

\definecolor{myblue}{rgb}{0,0,1}
\definecolor{myred}{rgb}{1,0,0}
\definecolor{mylightgray}{gray}{0.95}
\definecolor{myCite}{HTML}{1C4587}

\definecolor{highlightblue}{HTML}{185ABC}
\definecolor{cellHighlight}{HTML}{dbefff}

\usepackage{minitoc}

\noptcrule


\newcolumntype{L}[1]{>{\raggedright\let\newline\\\arraybackslash\hspace{0pt}}m{#1}}
\newcolumntype{C}[1]{>{\centering}m{#1}}

\newcolumntype{R}[1]{>{\raggedleft\let\newline\\\arraybackslash\hspace{0pt}}m{#1}}

\definecolor{ao}{rgb}{0.0, 0.0, 1.0}

\newcommand\vcent[1]{\vcenter{\hbox{#1}}}

\newcommand\loudspeaker[1][3]{\ensuremath{\vcent{\rule{.6ex}{.6ex}}\kern-.5ex
  \vcent{\scalebox{.6}[1]{\rotatebox[origin=center]{90}{$\blacktriangle$}}}
  \ifnum#1>0\relax\kern.05ex\vcent{\scalebox{.4}{\ttfamily)}}
  \ifnum#1>1\relax\kern-.4ex\vcent{\scalebox{.56}{\ttfamily)}}
  \ifnum#1>2\relax\kern-.55ex\vcent{\scalebox{.7}{\ttfamily)}}
  \fi\fi\fi}
}

\makeatletter
\renewcommand\subparagraph{
 \@startsection {subparagraph}{5}{\z@ }{3.25ex \@plus 1ex
 \@minus .2ex}{-1em}{\normalfont \normalsize \bfseries }}
\makeatother

\let\cite\citep
\hypersetup{
  citecolor = myCite,  
  linkcolor = myCite,   
  urlcolor  = myCite
}

\title{Counteracting Matthew Effect in Self-Improvement of LVLMs through Head-Tail Re-balancing}
\author{
    Xin Guo$^1$$^*$, Zhiheng Xi$^1$$^*$, Yiwen Ding$^1$, Yitao Zhai$^2$, Xiaowei Shi$^2$, \\ \textbf{Xunliang Cai$^2$, Tao Gui$^{1,3}$$^\dag$, Qi Zhang$^1$, Xuanjing Huang$^1$$^\dag$} \\
$^1$Fudan University \ $^2$Meituan \ $^3$Shanghai Innovation Institute \\
\texttt{\{tgui,xjhuang\}@fudan.edu.cn} 
}
\vspace{-50pt}
\begin{abstract}
Self-improvement has emerged as a mainstream paradigm for advancing the reasoning capabilities of large vision–language models (LVLMs), where models explore and learn from successful trajectories iteratively. However, we identify a critical issue during this process: the model excels at generating high-quality trajectories for simple queries (i.e., head data) but struggles with more complex ones (i.e., tail data). This leads to an imbalanced optimization that drives the model to prioritize simple reasoning skills, while hindering its ability to tackle more complex reasoning tasks. Over iterations, this imbalance becomes increasingly pronounced–a dynamic we term the ``Matthew effect''\footnote{Matthew effect is a sociological concept originally proposed by Robert K. Merton \cite{merton1968matthew}, which can be summarized as ``the rich get richer and the poor get poorer''.}–which ultimately hinders further model improvement and leads to performance bottlenecks. To counteract this challenge, we introduce four efficient strategies from two perspectives: distribution-reshaping and trajectory-resampling, to achieve head-tail re-balancing during the exploration-and-learning self-improvement process. Extensive experiments on Qwen2-VL-7B-Instruct and InternVL2.5-4B models across visual reasoning tasks demonstrate that our methods consistently improve visual reasoning capabilities, outperforming vanilla self-improvement by 3.86 points on average. 
\end{abstract}

\begin{document}

\doparttoc
\faketableofcontents

\begingroup
  \renewcommand\thefootnote{}
  \footnote{\textsuperscript{*}Equal contribution.
            \textsuperscript{\dag}Corresponding authors.}
  \addtocounter{footnote}{-1}
\endgroup

\vspace{-30pt}
\maketitle

\vspace{-20pt}
\section{Introduction}

\begin{wrapfigure}{r}{10cm}
\vspace{-21pt}
\centering
\includegraphics[width=0.85\linewidth]{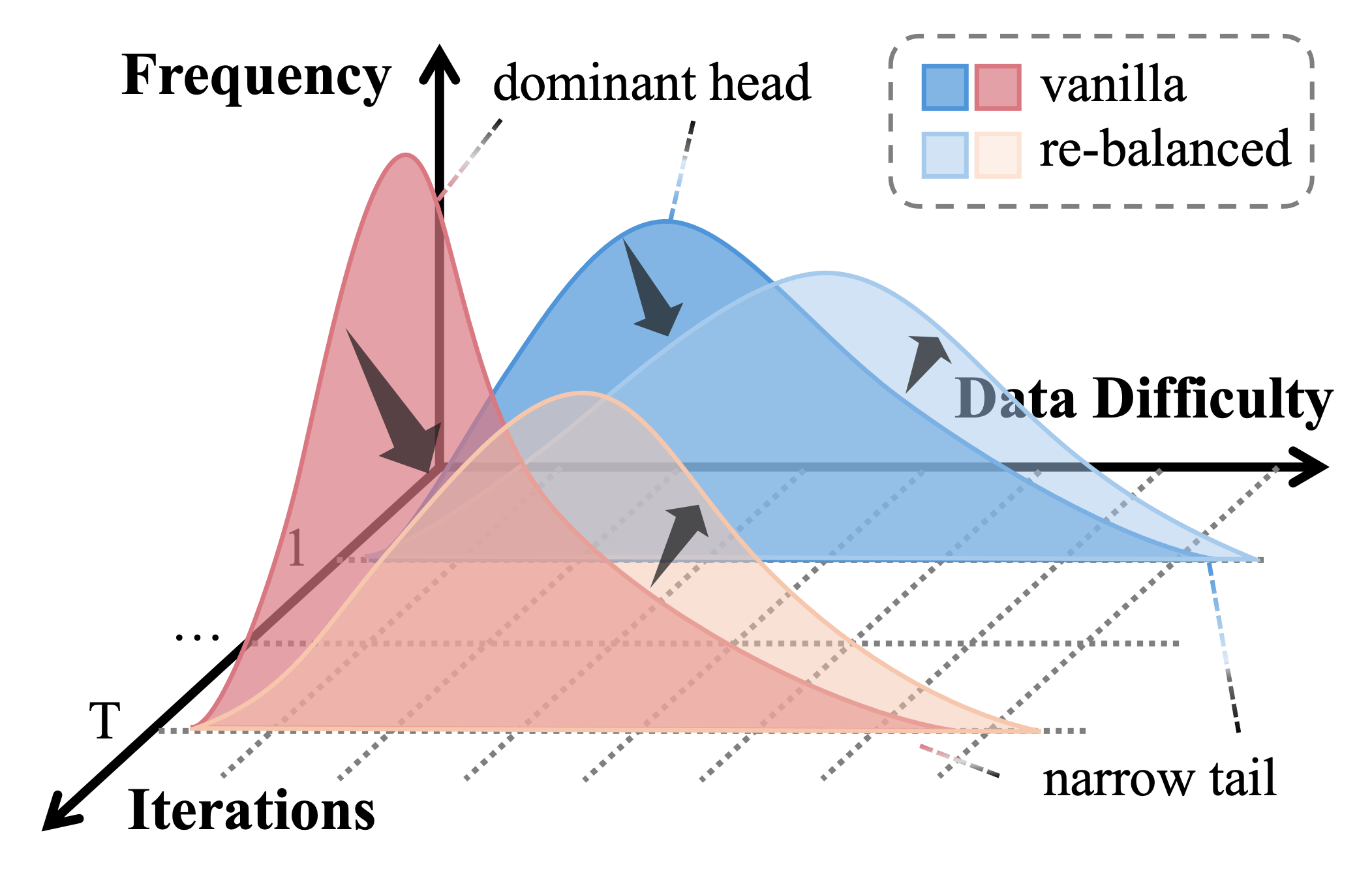}
\vspace{-3pt}
\caption{Matthew effect in self-improvement of LVLMs over iterations and our re-balanced solution. \textbf{Dark areas} illustrate the imbalanced distribution in vanilla self-improvement, where dominant head and narrow tail become more severe over iterations. \textbf{Light areas} depict re-balanced self-improvement–our methods for counteracting Matthew effect by reducing the head and augmenting the tail.}
\vspace{-20pt}
\label{fig:first}
\end{wrapfigure}

Large vision language models (LVLMs) have demonstrated impressive reasoning capabilities across complex multimodal tasks \cite{DBLP:journals/corr/abs-2502-13923, DBLP:journals/corr/abs-2504-10479}. While supervised fine-tuning (SFT) can further improve model performance, its effectiveness is limited by the current lack of sufficient large-scale, high-quality annotated datasets \cite{DBLP:conf/iclr/ZhangWJGZT0ZZG025, DBLP:journals/corr/abs-2504-05599}. 

In response, self-improvement has emerged as a promising alternative, enabling LVLMs to iteratively explore and learn from successful trajectories \cite{DBLP:conf/emnlp/0001GHW00023, DBLP:conf/nips/DengLYHSGZC024,  DBLP:conf/naacl/WangCWZZYZGBKHX25}. 
This paradigm not only eliminates reliance on manual annotations but also promotes better distribution alignment between self-generated and real-world data \cite{DBLP:conf/aaai/JiangCXLG25}. However, our preliminary experiments reveal that self-improvement with varying sampling numbers $K$ encounters performance bottleneck–and even decline–in visual reasoning scenarios (§\ref{sec:bottlenecks}). 

We delve into this process (§\ref{sec:imbalanced distribution}) and uncover a significant imbalance in the distribution of self-generated successful trajectories \cite{DBLP:conf/icml/DohmatobFYCK24, DBLP:conf/naacl/DingXHLZXCGZH25}. 
Specifically, we observe that simple samples overwhelmingly dominate the head, while challenging samples in the narrow tail remain underexplored. 
As illustrated in Figure \ref{fig:first}, this imbalance becomes increasingly severe as iterations progress: the dominant head expands further, whereas the narrow tail becomes more marginalized, which we term the ``Matthew effect'' \cite{merton1968matthew}. 
Further comparative analysis reveals the Matthew effect of self-generated data manifests in distinct ways; namely, a worsening imbalance across difficulty levels \cite{DBLP:conf/nips/TongZWWH24, DBLP:journals/corr/abs-2503-09029} and a trend toward increasingly shorter average response lengths \cite{DBLP:conf/acl/WangKMLSKH23}, ultimately leading to performance degradation.

To this end, we propose four effective strategies for head-tail re-balancing (§\ref{sec:methods}) inspired by \citet{DBLP:journals/corr/abs-2505-17652} and \citet{DBLP:conf/icml/XiCHJZHDLGWGSFZ24}. 
Intuitively, we first focus on \textit{distribution-reshaping} to better utilize existing sampled data. For head reduction, we introduce \textbf{threshold clipping}, which randomly truncates responses beyond a predefined threshold $L$ to limit successful trajectories per query.
% sets a threshold $L$ and randomly truncates responses to limit each query to at most $L$ successful trajectories. 
For tail augmentation, we propose \textbf{repeat-based padding}, which ensures equal frequency of all queries through repetition. 
Further, we augment tail data through \textit{trajectory-resampling} to boost data diversity, introducing \textbf{adaptive-weighted resampling}, which dynamically adjusts resampling weights based on fail rate \cite{DBLP:conf/nips/TongZWWH24}. However, this resampling strategy suffers from low efficiency. Therefore, we develop \textbf{guided resampling}, which efficiently explores tail data by initializing model reasoning from varying intermediate steps.

We conduct experiments on visual reasoning tasks across Qwen2-VL-7B-Instruct \cite{DBLP:journals/corr/abs-2409-12191} and InternVL2.5-4B \cite{DBLP:journals/corr/abs-2504-10479} models under the settings of sampling number $K=8$ and $K=16$, respectively (§\ref{sec:experiments}). 
The results demonstrate that our proposed methods effectively mitigate Matthew effect in visual self-improvement through head-tail re-balancing, achieving significant performance enhancements. 

Our contributions are summarized as follows:
\begin{itemize}[leftmargin=*]
\item We delve into the visual self-improvement process, revealing that the key challenge of performance bottlenecks is the imbalanced head-tail distribution and Matthew effect over iterations.
\item To address this, we propose a methodological framework for head-tail re-balancing, which integrates four strategies from distribution-reshaping and trajectory-resampling perspectives.
\item We conduct comprehensive experiments across different models and visual reasoning tasks, demonstrating the effectiveness of head-tail re-balancing in counteracting the Matthew effect within visual self-improvement.
\end{itemize}

\section{Related Work}
\subsection{Self-improvement in visual reasoning.}
LVLMs have demonstrated remarkable performance across various visual reasoning scenarios \cite{DBLP:journals/corr/abs-2502-13923, DBLP:journals/corr/abs-2504-10479}, where self-improvement approaches have been widely employed \cite{yang2023idea2img, DBLP:conf/nips/DengLYHSGZC024, DBLP:conf/naacl/WangCWZZYZGBKHX25}. Among these, self-critic \cite{DBLP:conf/naacl/WangCWZZYZGBKHX25} and self-correction \cite{DBLP:journals/corr/abs-2505-22651, Wu_2025_CVPR, DBLP:conf/naacl/ChengLXZZL25} 
emerge as prevalent optimization strategies.
Prior work typically relies on separate critic models for error detection and correction, which requires substantial additional resources \cite{DBLP:journals/corr/abs-2502-13383, Xiong_2025_CVPR, Zhang_2025_CVPR}. In contrast, \citet{DBLP:conf/naacl/WangCWZZYZGBKHX25} introduce a unified model that simultaneously generates responses and performs self-critic to refine. Similarly motivated, \citet{DBLP:journals/corr/abs-2505-22651} propose Sherlock to selectively revise erroneous segments in reasoning trajectories.
Moreover, to enhance data efficiency, many studies employ direct preference optimization (DPO) in self-improvement \cite{DBLP:conf/nips/DengLYHSGZC024, DBLP:conf/naacl/WangCWZZYZGBKHX25, DBLP:journals/corr/abs-2506-02708}.
However, these works lack a thorough exploration of the iterative process. 

In this paper, we focus on the essence of vanilla self-improvement and propose targeted and effective strategies to enhance its performance.

\subsection{Distribution bias in self-generated data.}
Previous work has identified biases sampling and head-tail imbalance in self-improvement sampling process \cite{DBLP:journals/nature/ShumailovSZPAG24, DBLP:conf/icml/DohmatobFYCK24} and addressed them by increasing the proportion of difficult queries through adaptive-weighted \cite{DBLP:conf/nips/TongZWWH24, DBLP:journals/corr/abs-2503-09029, DBLP:journals/corr/abs-2505-17652} and guided \cite{DBLP:conf/naacl/DingXHLZXCGZH25} sampling methods. However, these studies primarily focus on text-based reasoning while lacking thorough investigation into visual scenarios. While SynthRL \cite{DBLP:journals/corr/abs-2506-02096} enhances the distribution balance by rewriting queries on visual reasoning tasks, it does not delve into the iterative process.

Instead, we conduct an in-depth investigation into the observable performance and underlying properties of LVLMs during self-improvement, revealing similar distribution bias. To this end, we introduce four re-balancing strategies, effectively mitigating the Matthew effect and achieving significant performance improvements.

\section{``Matthew Effect'' in Self-improvement}\label{sec:analysis}
While existing research has explored self-generated data, the self-improvement process in visual reasoning scenarios remains underexplored. Therefore, we delve into this process to uncover its intrinsic properties.

\subsection{Formulating Self-improvement}
In this paper, we formulate the vanilla self-improvement as follows.
Given a model $\mathcal{M}_{\textnormal{base}}$, a training dataset $\mathcal{D}_{\textnormal{base}}=\{(q_i, a_i)\}_{i=1}^N$ where $q_i$ represents the query, $a_i$ denotes the ground-truth answer, and $N$ is the dataset size, we define the initial model as $\mathcal{M}_0=\mathcal{M}_{\textnormal{base}}$ and the number of iterations as $T$.
At each iteration $t\in [1, T]$, the self-improvement sequentially performs three key stages: exploration, filtering, and learning.

\paragraph{Exploration.} 
At iteration $t$, for each query $q_i\in \mathcal{D}_{\textnormal{base}}$, the model $\mathcal{M}_{t-1}$ generates $K$ different responses $\{\hat{r}_{i, k}^{(t)}\}_{k=1}^{K}$, where $\hat{r}_{i, k}^{(t)} \sim \mathcal{M}_{t-1}(\cdot | q_i)$. Therefore, the sampled dataset is represented as 
\begin{equation*}\mathcal{D}_{\textnormal{sample}}^{(t)}=\{(q_i, \hat{r}_{i, k}^{(t)})\ |\ 1 \leq i \leq N, 1 \leq k \leq K\}.\end{equation*}

\paragraph{Filtering.}
To obtain high-quality training data, we define a binary reward function as 
\begin{equation*}rf(q_i, a_i, \hat{a}_{i,k}^{(t)})=\begin{cases}
0, & \text{ if } \hat{a}_{i,k}^{(t)}\ne a_i \\
1, & \text{ if } \hat{a}_{i,k}^{(t)}=a_i
\end{cases},\end{equation*}
where $\hat{a}_{i,k}^{(t)}$ is the answer extracted from response $\hat{r}_{i,k}^{(t)}$. Using it, we obtain the filtered dataset: \begin{equation*}\mathcal{D}_{\textnormal{filter}}^{(t)}=\{(q_i, \hat{r}_{i,k}^{(t)})\in \mathcal{D}_{\textnormal{sample}}^{(t)}\ |\ rf(q_i, a_i, \hat{a}_{i,k}^{(t)})=1\}.\end{equation*}

\paragraph{Learning.} At iteration $t$, the model $\mathcal{M}_t$ is trained by fine-tuning the base model $\mathcal{M}_{\textnormal{base}}$ on the filtered dataset $\mathcal{D}_{\textnormal{train}}^{(t)}=\mathcal{D}_{\textnormal{filter}}^{(t)}$. The objective is to minimize the negative log-likelihood:
\begin{equation*}\mathcal{L}_{\textnormal{SFT}}^{(t)} = -\mathbb{E}_{(q, r) \sim \mathcal{D}_{\textnormal{train}}^{(t)}} \sum_{j=1}^{|r|} \log P(r_j | q, r_{<j}; \mathcal{M}_t).\end{equation*}

Through $T$ iterations of this process, the model continuously generates higher-quality responses, which in turn enhances its performance in the next iterations, thereby achieving self-improvement.

\subsection{Plateaus in Varying Sampling Numbers}\label{sec:bottlenecks}
\begin{figure*}[t]
  \centering
    \begin{subfigure}[t]{0.27\linewidth}
        \includegraphics[width=0.9\linewidth]{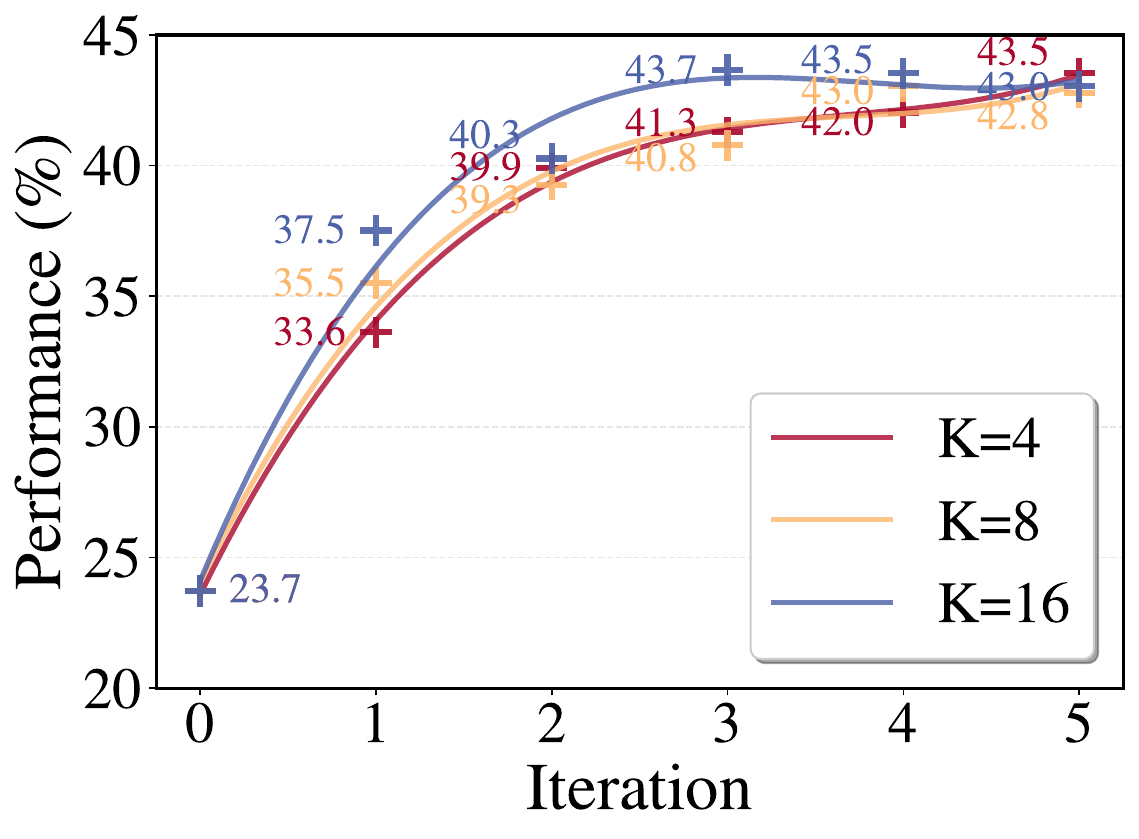} \caption{Performance bottlenecks.}
        \label{fig:performance_of_different_k}
    \end{subfigure}
    % \hfil
    \begin{subfigure}[t]{0.37\linewidth}
        \includegraphics[width=0.92\linewidth]{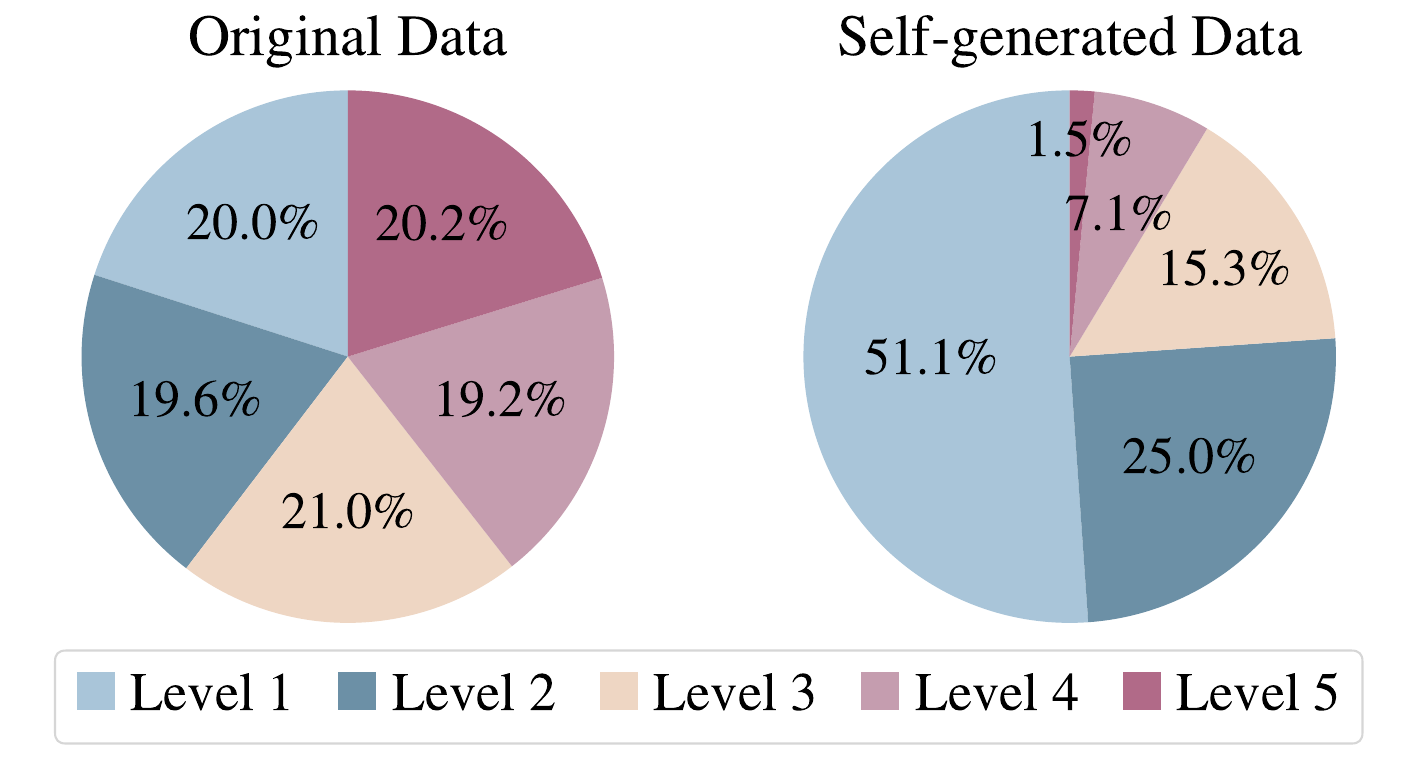} \caption{Distribution of difficulty level.}
        \label{fig:ratio_of_different_level}
    \end{subfigure}
    % \hfil
    \begin{subfigure}[t]{0.34\linewidth}
        \includegraphics[width=0.9\linewidth]{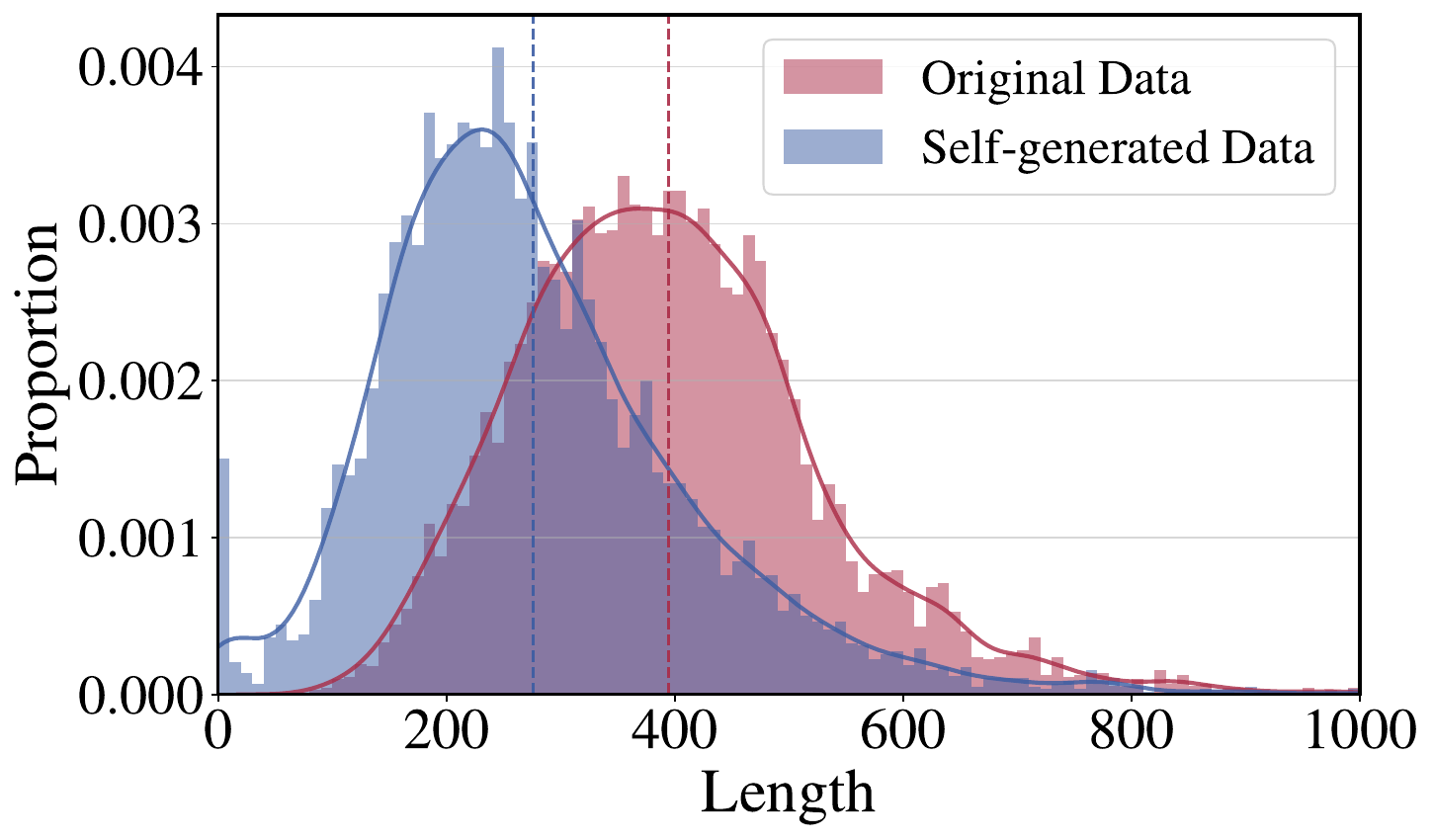}
        \caption{Distribution of response length.}
        \label{fig:ratio_of_different_length}
    \end{subfigure}
    \caption{Performance bottlenecks and distribution characteristics in self-improvement. \textbf{(a)} Phenomenon of performance bottlenecks under different sampling numbers $K$. \textbf{(b)} Comparison of difficulty level distributions between original and self-generated data, ranging from level 1 (easiest) to level 5 (most difficult). \textbf{(c)} Differences in response length distributions between original and self-generated data, with dashed lines indicating mean values.}
\end{figure*}

First, to reveal the impact of sampling number $K$, we conduct experiments over five self-improvement iterations with varying value of $K$. Figure \ref{fig:performance_of_different_k} uncovers two key observations: (1) Different $K$ exhibits similar performance trends during self-improvement: notable improvements in early iterations, rapid convergence to performance bottlenecks, and even declining trends (e.g., $K=16$) in later iterations; (2) Sampling number plays a dominant role in the first iteration, where higher sampling number leads to better performance; while in later iterations, sampling number has virtually no impact on performance. For instance, at iteration $1$, the $K=16$ model outperforms the $K=4$ model by $3.89$ points on the test set. While at iteration $5$, the $K=16$ model even performs $0.39$ points worse than the $K=4$ model.

\subsection{Imbalance in Self-improvement}\label{sec:imbalanced distribution}
To figure out the bottlenecks of self-improvement, we then analyze the distribution of self-generated data \cite{DBLP:conf/nips/TongZWWH24, DBLP:conf/naacl/DingXHLZXCGZH25}.

\paragraph{Differences between original and self-generated data.}
To characterize the properties of self-generated data, we first compare it with the original data from two dimensions: \textit{difficulty} and \textit{length}.
\begin{itemize}[leftmargin=*]
\item \textit{\textbf{Imbalanced difficulty distribution.}} As shown in Figure \ref{fig:ratio_of_different_level}, the difficulty distribution (see Appendix \ref{appendix:A} for difficulty level categorization) differs markedly between the original and self-generated data, revealing a ``dominant head and narrow tail'' pattern.
% Figure \ref{fig:ratio_of_different_level} illustrates the distribution of different difficulty levels across both original and self-generated data, revealing the phenomenon of dominant head and narrow tail. 
While original data exhibits balanced difficulty distribution, self-generated data suffers from severe imbalance, with easy samples (level 1) comprising 51.1\% of the total and difficult samples (level 5) nearly absent.

\item \textit{\textbf{Shorter average response length.}} 
As revealed in Figure \ref{fig:ratio_of_different_length}, self-generated responses are significantly shorter than original ones (averaging 277 tokens vs. 395 tokens), with some extremely short responses (\textless 10 tokens) lacking CoT reasoning. This suggests that self-generated data is prone to producing abbreviated reasoning processes, including instances that deliver conclusions directly even under CoT prompting.
\end{itemize}

\begin{figure*}[t]
\centering
  \begin{subfigure}[t]{0.36\textwidth}
  \centering
  \includegraphics[width=0.99\linewidth]{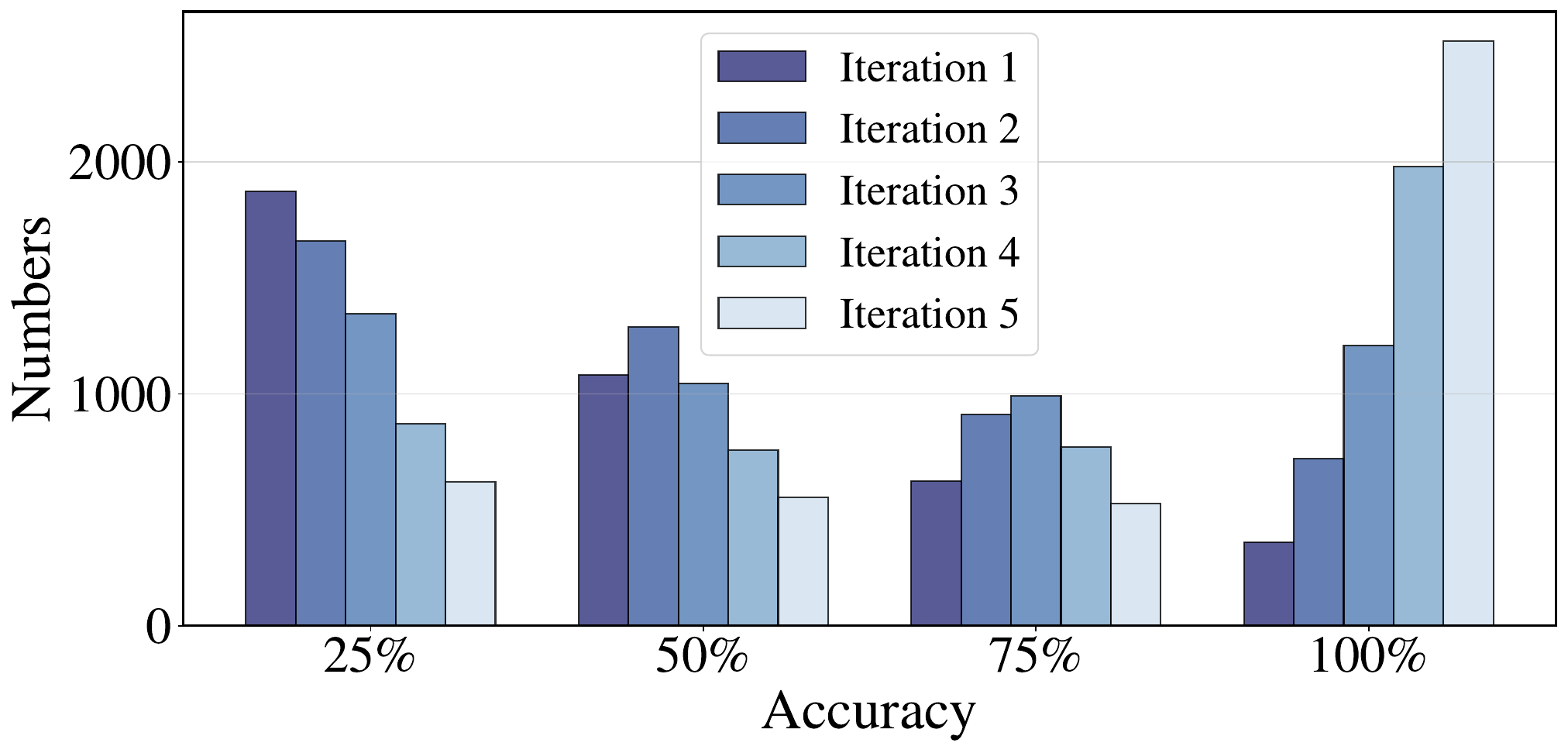}
  \caption{Data with different accuracy.}
  \label{fig:distribution_k4_all}
  \end{subfigure}
  \hfil
  \begin{subfigure}[t]{0.33\textwidth}
  \centering
  \includegraphics[width=0.9\linewidth]{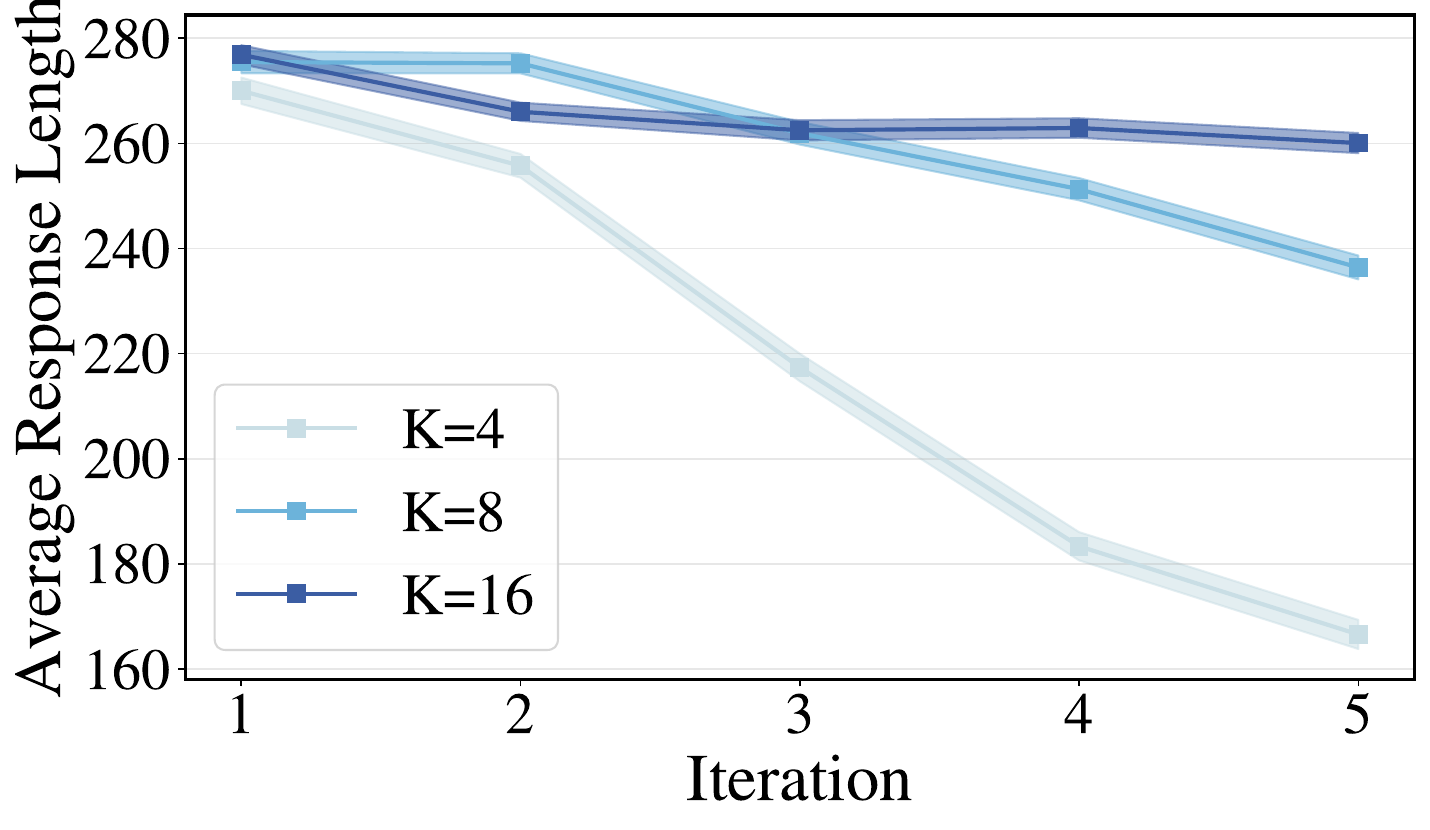}
  \caption{Length with different $K$.}
  \label{fig:length_different_k}
  \end{subfigure}
  \hfil
  \begin{subfigure}[t]{0.28\textwidth}
  \centering
  \includegraphics[width=0.86\linewidth]{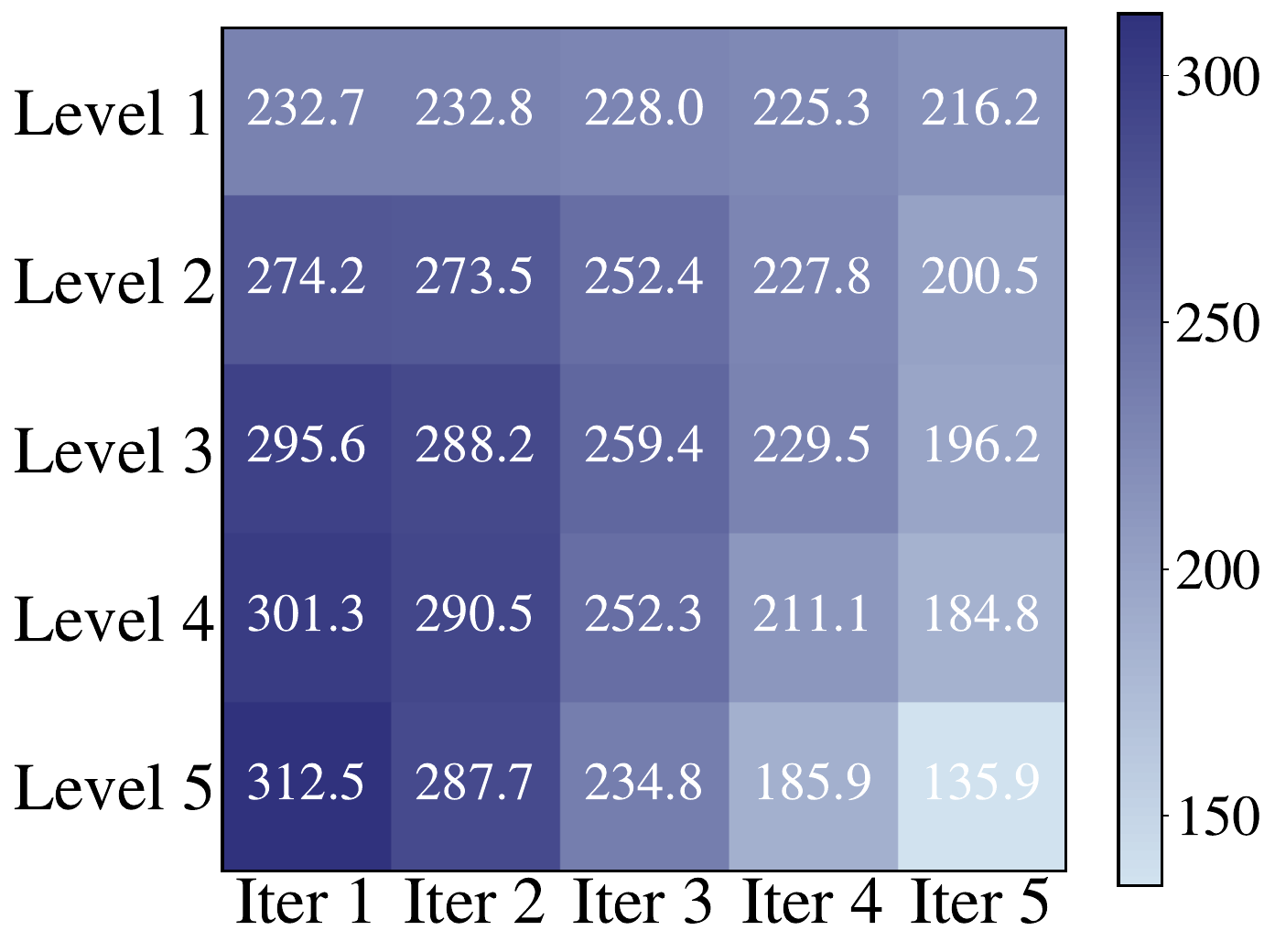}
  \caption{Length with different level.}
  \label{fig:length_shifts_k8}
  \end{subfigure}
  \caption{Matthew effect over iterations. \textbf{(a)} Matthew effect in the distribution of data in $\mathcal{D}_{\textnormal{filter}}$ with different accuracy under the setting of $K=4$. \textbf{(b)} Trends in average response length (i.e., number of tokens) across iterations under different sampling numbers $K$. \textbf{(c)} Matthew effect in average response length for data of different difficulty levels across iterations under the setting of $K=8$, where difficulty increases progressively from level 1 to level 5.}
\end{figure*}

\paragraph{Matthew effect over iterations.}
Further, we investigate distribution shifts during the iterative process. 
To begin with, we analyze how queries with different accuracy are distributed in self-generated data. 
Results in Figure \ref{fig:distribution_k4_all} indicate that throughout iterations, well-mastered data (with 100\% accuracy) occupies an increasing proportion, while poorly-performed data (with 25\% accuracy) is gradually squeezed out of the training dataset $\mathcal{D}_{\textnormal{train}}$–a phenomenon we term the ``Matthew effect''. Given that high-accuracy queries account for more samples, the diminishing proportion of difficult tail becomes even more severe, limiting the self-improvement performance ceilings.

Next, we analyze response length changes during the iterative process. As shown in Figure \ref{fig:length_different_k}, average response length consistently declines across iterations, while higher sampling number mitigates this degradation. To delve deeper, we compare average response length across various difficulty levels over iterations. Figure \ref{fig:length_shifts_k8} reveals three key observations: (1) Simple data (level 1) maintains relatively stable response length during iterations, showing minimal degradation. (2) Difficult data (level 5) suffers the most severe length degradation, with a dramatic reduction of 56.5\%. (3) At iteration 1, the higher the difficulty level, the longer the response length; however, this trend completely reverses by the fifth iteration, with difficult data generating the shortest responses of merely 136 tokens. These findings suggest that Matthew effect manifests in response length as well–simple data requiring shorter responses maintains appropriate length, whereas difficult data requiring longer responses undergoes significant degradation.

\section{Methodology}\label{sec:methods}
To alleviate the phenomenon of dominant head and narrow tail, we propose four effective re-balanced strategies from two perspectives: distribution-reshaping and trajectory-resampling.

Regarding distribution-reshaping, motivated by under-sampling and over-sampling in machine learning \cite{2020Machine}, we propose two intuitive strategies. On one hand, we reduce the quantity of head to increase the proportion of tail samples, introducing the \textbf{Threshold Clipping (TC)} strategy. On the other hand, we directly augment the number of tail through repetition, proposing the \textbf{Repeated-based Padding (RP)} strategy.

From trajectory-resampling perspective, we also propose two strategies. The first is \textbf{Adaptive-weighted Resampling (AR)}, which dynamically adjusts resampling weights based on fail rate. 
The second is \textbf{Guided Resampling (GR)}, which initializes model exploration from varying intermediate reasoning steps \cite{DBLP:conf/icml/XiCHJZHDLGWGSFZ24, DBLP:conf/naacl/DingXHLZXCGZH25}.

\subsection{Distribution Reshaping}
\paragraph{Threshold clipping.} Adopting the philosophy of ``less is more'', we propose threshold clipping, which increases the proportion of tail by reducing the number of head. 

Specifically, threshold clipping sets a threshold $L$ and randomly truncates responses to limit each query to at most $L$ correct responses. Formally, at iteration $t$, instead of using the filtered dataset directly, we train the base model $\mathcal{M}_{\textnormal{base}}$ on the following dataset: $$\mathcal{D}_{\textnormal{train-TC}}^{(t)}=\{(q_i, \hat{r}_{i,k}^{(t)})\in \mathcal{D}_{\textnormal{filter}}^{(t)}\ |\ k\le L\}.$$

\paragraph{Repeat-based padding.}
Moreover, we consider increasing the quantity of tail directly and introduce repeat-based padding to enforce balanced data distribution. 

Specifically, repeat-based padding ensures that all queries appear with equal frequency $K$ in the next training dataset. To achieve this, queries with insufficient correct samples are padded through repetition. Formally, at iteration $t$, the training dataset is expressed as follows: 
$$\mathcal{D}_{\textnormal{train-RP}}^{(t)}=\{(q_i, \hat{r}_{i, k \bmod k_{i}}^{(t)})\in\mathcal{D}_{\textnormal{filter}}\ |\ 1 \leq k \leq K\},$$
where $k_i$ is the number of $\hat{r}_{i}$, that is, the number of correct responses out of $K$ sampling for $q_i$.

\subsection{Trajectory Resampling}
\paragraph{Adaptive-weighted resampling.}
However, merely reshaping data distribution yields limited benefits. As excessive duplication may reduce data diversity and trigger overfitting, we additionally employ resampling strategies to enhance the proportion of tail. Drawing inspiration from the ``pass rate'' \cite{DBLP:journals/corr/abs-2501-12599} and ``fail rate'' \cite{DBLP:conf/nips/TongZWWH24} metrics, we propose adaptive-weighted resampling, which dynamically adjusts the resampling weights for each query based on its fail rate. 

Specifically, for a query with $k_i$ successful trajectories out of $K$ samples, we perform $K-k_i$ additional resampling operations. 
This hierarchical resampling assigns more weight to tail data, thereby increasing their proportion. Formally, at iteration $t$, model $\mathcal{M}_{t-1}$ generates $K-k_i$ new responses $\{\check{r}_{i, k}^{(t)}\}_{k=1}^{K-k_i}$ for each query $q_i\in \mathcal{D}_{\textnormal{base}}$, where $\check{r}_{i, k}^{(t)}\sim \mathcal{M}_{t-1}(\cdot | q_i)$. Then we obtain the resampled and refiltered datasets: 

$$\mathcal{D}_{\textnormal{resample-AR}}^{(t)}=\{(q_i, \check{r}_{i, k}^{(t)})\ |\ 1 \leq i \leq N, 1 \leq k \leq K-k_i\}$$

$$\mathcal{D}_{\textnormal{refilter-AR}}^{(t)}=\{(q_i, \check{r}_{i,k}^{(t)})\in \mathcal{D}_{\textnormal{resample-AR}}^{(t)}\ |\ rf(q_i, a_i, \check{a}_{i,k}^{(t)})=1\},$$
and finally merge into the training dataset: 
$$\mathcal{D}_{\textnormal{train-AR}}^{(t)}=\mathcal{D}_{\textnormal{filter}}^{(t)}\cup\mathcal{D}_{\textnormal{refilter-AR}}^{(t)}.$$

\paragraph{Guided resampling.}
Nevertheless, adaptive-weighted resampling is essentially brute-force sampling with limited efficiency improvements.
To this end, we propose guided resampling–a novel resampling strategy that initializes model exploration from various intermediate reasoning steps. 

Specifically, the model exploits guided signals to achieve efficient resampling, starting its reasoning process from different intermediate reasoning steps. This strategy enables the model to navigate toward promising trajectories within a vast exploration space, while facilitating progressive learning of complex reasoning. Formally, for each successful trajectory $\hat{r}_i\in\mathcal{D}_{\textnormal{filter}}^{(t)}$, we decompose it into $S$ steps $\hat{r}_{i(1)}, \cdots, \hat{r}_{i(S)}$. Generated by $\check{r}_{i(s)}\sim\mathcal{M}_{t-1}(\cdot|q_i, \hat{r}_{i(<s)})$, the resample dataset is expressed as:
$$\mathcal{D}_{\textnormal{resample-GR}}^{(t)}=\{(q_i, \hat{r}_{i(<s), k}^{(t)}, \check{r}_{i(s), k}^{(t)})\ |\ k_i<L, 1 \leq s \leq S\}.$$
Similar to adaptive-weighted resampling, we obtain the refiltered and training dataset as $\mathcal{D}_{\textnormal{refilter-GR}}^{(t)}$ and $\mathcal{D}_{\textnormal{train-GR}}^{(t)}$.

\section{Experiments}\label{sec:experiments}
\subsection{Experimental Setups}
\paragraph{Datasets.}
We adopt MMPR \cite{DBLP:journals/corr/abs-2411-10442}–a multimodal reasoning dataset derived from multiple sources–as our primary dataset. From it, we randomly extract 7,980 mathematical reasoning samples \cite{DBLP:conf/acl/LuGJQHLZ20, DBLP:conf/coling/CaoX22, DBLP:conf/emnlp/SeoHFEM15} to construct a curated subset, MMPR-mini, with details presented in Appendix \ref{appendix:A}. 
For out-of-domain (OOD) evaluation, we further utilize MathVerse \cite{DBLP:conf/eccv/ZhangJZLGQZLCQGL24} and We-Math \cite{DBLP:journals/corr/abs-2407-01284} datasets. 

\paragraph{Models.}
We employ two widely-used LVLMs as our base models: Qwen2-VL-7B-Instruct \cite{DBLP:journals/corr/abs-2409-12191} and InternVL2.5-4B \cite{DBLP:journals/corr/abs-2504-10479}. All analytical experiments in Section \ref{sec:analysis} are conducted on Qwen2-VL-7B-Instruct.
Following the paradigm of \citet{zelikman2024star}, we initialize the from the base model instead of a further SFT model, and restart training from this base model at each iteration.

\paragraph{Implementation details.}
All experiments are conducted on eight A100-80GB GPUs using SWIFT \cite{DBLP:conf/aaai/ZhaoHHWMZJWAWZC25} framework for training and vLLM \cite{DBLP:conf/sosp/KwonLZ0ZY0ZS23} framework for sampling and testing. We set the number of self-improvement iterations $T=5$ and focus on the sampling number $K=8$ and $K=16$. For training, we use a learning rate of $3\times10^{-5}$ and train for $1$ epoch to avoid overfitting. For sampling, we set the temperature to $0.7$ with maximum $4096$ new tokens. While for testing, the temperature is set to $0$. Additionally, we configure method-specific parameters, including threshold $L=4$ for TC and number of steps $S=4$ for AR.

\subsection{Main Results}
\begin{table}[t]
\centering 
\resizebox{\linewidth}{!}{
\begin{tabular}{ccccccccccccccc}
\toprule
\multirow{3}{*}{\textbf{Models}} & \multirow{3}{*}{\textbf{Sampling Num.}} & \multirow{3}{*}{\textbf{Method}} & \multicolumn{2}{c}{\textbf{In-domain}} & \multicolumn{4}{c}{\textbf{Out-of-domain}}      & \multicolumn{2}{c}{\multirow{2}{*}{\textbf{Avg.}}} \\ 
\cmidrule(r){6-9}
                                      &             &            & \multicolumn{2}{c}{\textbf{MMPR-mini}} & \multicolumn{2}{c}{\textbf{MathVerse}} & \multicolumn{2}{c}{\textbf{We-Math}} &                          \\
& & & final & opt. & final & opt. & final & opt. & final & opt.  \\
\midrule
\multirow{11}{*}{Qwen2-VL-7B} & \multicolumn{2}{c}{w/o SI} & \multicolumn{2}{c}{$23.71$} & \multicolumn{2}{c}{$22.46$} & \multicolumn{2}{c}{$47.36$} & \multicolumn{2}{c}{$31.18$} \\
\cdashline{2-11}
& \multirow{5}{*}{$K=8$} & Vanilla SI & $42.79$ & $43.04$ & $27.79$ & $28.93$ & $52.36$ & $\underline{52.93}$ & $40.98$ & $41.31$ \\ 
                                      & & TC & $44.67$ & $44.67$ & $28.55$ & $\underline{29.06}$ & $52.13$ & $52.76$ & $41.78$ & $41.78$ \\  
                                      & & RP & $\underline{45.67}$ & $\mathbf{46.42}$ & $\mathbf{30.08}$ & $\mathbf{30.08}$ & $52.59$ & $52.59$ & $\mathbf{42.78}$ & $\mathbf{42.78}$ \\  
                                      & & AR & $\underline{45.67}$ & $45.67$ & $\underline{28.93}$ & $28.93$ & $\underline{52.82}$ & $52.82$ & $\underline{42.47}$ & $\underline{42.47}$ \\ 
                                      & & GR & $\mathbf{45.80}$ & $\underline{45.80}$ & $27.28$ & $28.30$ & $\mathbf{53.91}$ & $\mathbf{53.91}$ & $42.33$ & $42.33$ \\
\cdashline{2-11}
& \multirow{5}{*}{$K=16$} & Vanilla SI & $43.04$ & $43.66$ & $28.30$ & $28.30$ & $52.76$ & $52.76$ & $41.36$ & $41.36$ \\ 
                                      & & TC & $45.92$ & $45.92$ & $\underline{28.68}$ & $28.68$ & $52.70$ & $52.76$ & $\underline{42.43}$ & $42.43$ \\  
                                      & & RP & $\underline{47.05}$ & $\underline{47.05}$ & $28.30$ & $\mathbf{30.33}$ & $51.49$ & $52.70$ & $42.28$ & $\underline{42.56}$\\  
                                      & & AR & $43.41$ & $43.41$ & $28.55$ & $28.93$ & $\underline{53.91}$ & $\underline{53.91}$ & $41.96$ & $41.96$ \\ 
                                      & & GR & $\mathbf{47.68}$ & $\mathbf{47.68}$ & $\mathbf{29.95}$ & $\underline{29.95}$ & $\mathbf{54.20}$ & $\mathbf{54.20}$ & $\mathbf{43.94}$ & $\mathbf{43.94}$ \\
\midrule
\multirow{11}{*}{InternVL2.5-4B} & \multicolumn{2}{c}{w/o SI} & \multicolumn{2}{c}{$47.55$} & \multicolumn{2}{c}{$29.31$} & \multicolumn{2}{c}{$48.10$} & \multicolumn{2}{c}{$41.66$} \\
\cdashline{2-11}
& \multirow{5}{*}{$K=8$} & Vanilla SI & $64.62$ & $64.62$ & $29.95$ & $\mathbf{33.12}$ & $50.40$ & $52.59$ & $48.32$ & $48.80$ \\ 
                                      & & TC & $63.99$ & $63.99$ & $\underline{32.36}$ & $\mathbf{33.12}$ & $\underline{51.72}$ & $\underline{53.28}$ & $49.36$ & $49.77$ \\  
                                      & & RP & $\underline{67.13}$ & $\mathbf{68.13}$ & $31.98$ & $31.98$ & $50.86$ & $52.87$ & $\underline{49.99}$ & $\underline{50.54}$ \\  
                                      & & AR & $66.25$ & $66.25$ & $30.33$ & $31.85$ & $50.75$ & $52.41$ & $49.11$ & $49.11$ \\ 
                                      & & GR & $\mathbf{67.50}$ & $\underline{67.50}$ & $\mathbf{33.12}$ & $\mathbf{33.12}$ & $\mathbf{51.90}$ & $\mathbf{54.20}$ & $\mathbf{50.84}$ & $\mathbf{50.84}$ \\
\cdashline{2-15}
& \multirow{5}{*}{$K=16$} & Vanilla SI & $66.75$ & $66.75$ & $\underline{31.60}$ & $\underline{33.63}$ & $50.46$ & $52.30$ & $49.60$ & $\underline{50.57}$ \\ 
                                      & & TC & $64.74$ & $64.74$ & $30.33$ & $33.38$ & $48.74$ & $50.57$ & $47.94$ & $48.99$ \\  
                                      & & RP & $\mathbf{71.64}$ & $\mathbf{71.64}$ & $\mathbf{32.61}$ & $\mathbf{34.14}$ & $51.32$ & $\mathbf{53.33}$ & $\mathbf{51.86}$ & $\mathbf{51.86}$ \\  
                                      & & AR & $64.99$ & $64.99$ & $30.58$ & $33.25$ & $\mathbf{51.67}$ & $51.78$ & $49.08$ & $49.46$ \\ 
                                      & & GR & $\underline{67.63}$ & $\underline{67.63}$ & $31.09$ & $31.98$ & $\underline{51.38}$ & $\underline{52.70}$ & $\underline{50.03}$ & $50.03$ \\
\bottomrule                         
\end{tabular}
}
\caption{Main results. The best result for each setting is in \textbf{bold}, while the second-best is marked with \underline{underline}. ``Num.'' refers to number. SI, TC, RP, AR, and GR denote self-improvement, threshold clipping, repeat-based padding, adaptive-weighted resampling, and guided resampling, respectively. ``Final'' indicates the final performance, and ``opt.'' indicates the optimal performance across all iterations. }
\vspace{-5pt}
\label{tab:main}
\end{table}

The main results are presented in Table \ref{tab:main}, including the final and optimal performance. 
Overall, our findings are as follows:

\paragraph{Scaling the sampling number $\mathbf{K}$ shows poor cost-efficiency in self-improvement.}
While vanilla self-improvement yields substantial gains over the base model–for instance, improving the average performance of Qwen2-VL-7B-Instruct by $10.13$ points at a sampling number of $K=8$–further increasing the sampling number proves ineffective. Compared to $K=8$, the optimal average performance at $K=16$ improves by merely $0.05$ points. Given the doubled computational cost, such a marginal gain is not cost-effective. These findings indicate that blindly scaling the sampling number through brute-force methods fails to provide critical breakthrough in self-improvement.

\paragraph{Head-tail re-balancing improves the performance of self-improvement across various models and datasets.}
For more efficient enhancement, re-balancing the distribution of head and tail data throughout self-improvement achieves significant performance improvements across varying models and datasets, particularly with RP and GR strategies. For example, with Qwen2-VL-7B-Instruct model at $K=16$, RP outperforms vanilla self-improvement by $3.39$ points on the in-domain test set and $1.20$ points on average, while GR achieves improvements of $4.02$ and $2.58$ points, respectively.

Moreover, comparison between final and optimal performance shows that vanilla self-improvement frequently exhibits suboptimal results in the final iteration relative to its peak performance, reflecting performance bottlenecks during training. In contrast, our re-balancing strategies, especially GR, consistently reach optimal performance in the final iteration, demonstrating greater stability and potential for further improvement.

\begin{wrapfigure}{r}{9cm}
\centering
\vspace{-2pt}
\includegraphics[width=\linewidth]{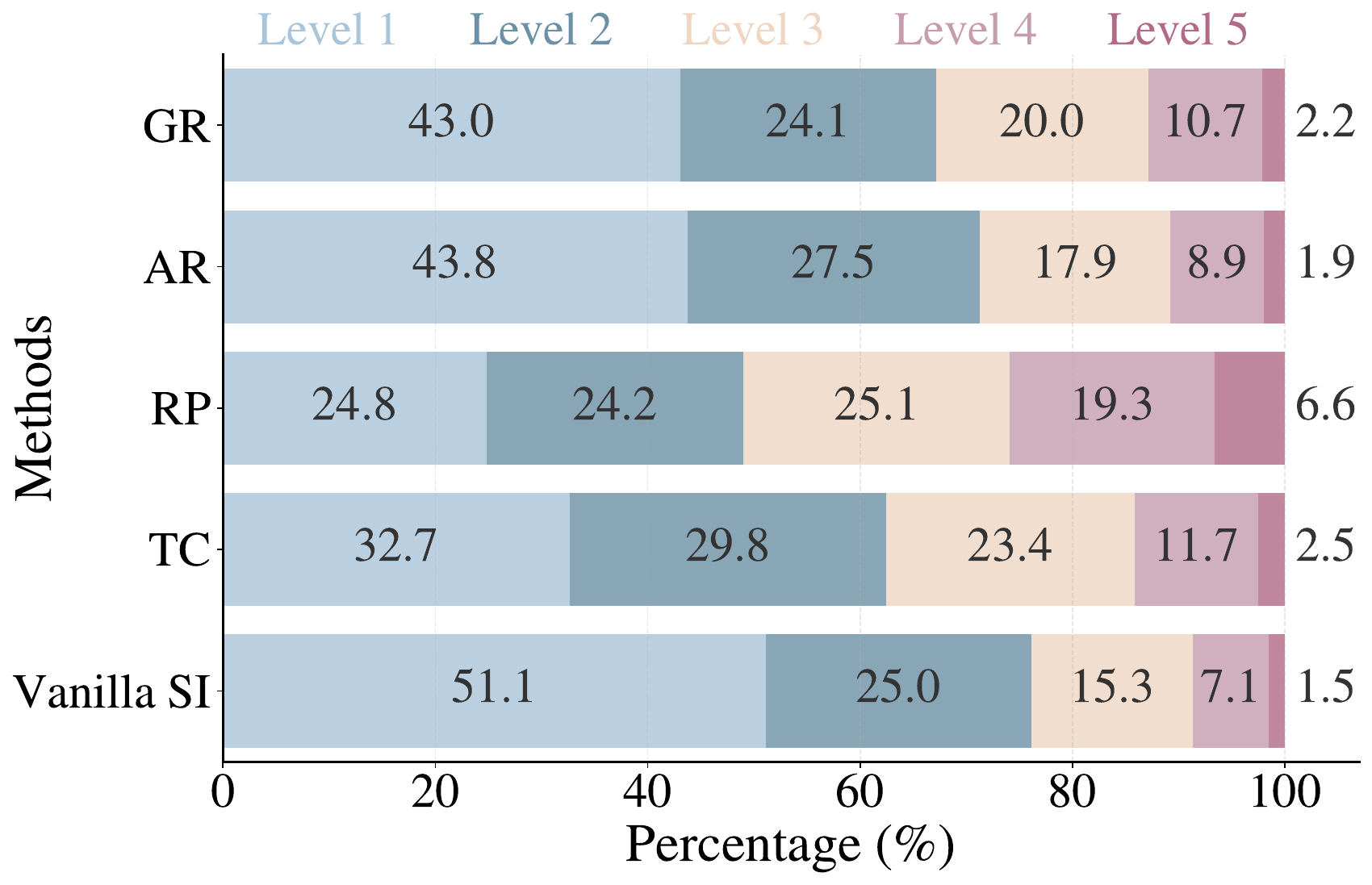}
\vspace{-12pt}
\caption{Data distribution of difficulty levels (1=easiest, 5=most difficult) in successful trajectories under different strategies with Qwen2-VL-7B-Instruct at $K=16$.}
\label{fig:strategies}
\end{wrapfigure}

\paragraph{Head-tail re-balancing mitigates the Matthew effect in self-improvement.}
As shown in Figure \ref{fig:strategies}, our proposed strategies effectively alleviate the imbalanced difficulty distribution in successful trajectories. Among them, distribution-reshaping methods exhibit superior mitigation on Qwen2-VL-7B-Instruct at $K=16$, with RP reducing head data proportion from 51.1\% to 24.8\% and boosting tail data from 1.5\% to 6.6\%. In contrast, AR shows limited tail data augmentation due to inefficient sampling in vast solution spaces, leading to modest performance gains. Conversely, GR focuses more on enhancing tail sample coverage and proportion. Despite slightly less mitigation than TC and RP, GR provides step-by-step guidance for tail samples, enables better mastery of complex reasoning trajectories, and yields substantial performance improvements from $41.36$ to $43.94$.
Additional results of Matthew effect mitigation on InternVL2.5-4B and other sampling settings are provided in Appendix \ref{appendix:D}.

\section{Discussion}\label{sec:discussion}
\subsection{Ablation Study}

\begin{wraptable}{r}{8.9cm}
\centering 
\vspace{-10pt}
\resizebox{\linewidth}{!}{
\begin{tabular}{lcccc}
\toprule
\textbf{Method} & \textbf{MMPR} & \textbf{MathVerse} & \textbf{We-Math} & \textbf{Avg.} \\
\midrule
TC ($L=2$) & $43.91$ & $27.66$ & $53.10$ & $41.56$\\  
\textbf{TC ($\bm{L=4}$)} & $45.92$ & $28.68$ & $52.70$ & $\mathbf{42.43}$ \\  
TC ($L=8$) & $43.66$ & $29.44$ & $52.47$ & $41.86$\\
\cdashline{2-5}
HC & $41.78$ & $28.93$ & $53.10$ & $41.27$ \\
RI & $45.04$ & $28.81$ & $51.26$ & $41.71$ \\
\textbf{RP} & $47.05$ & $28.30$ & $51.49$ & $\mathbf{42.28}$ \\  
\midrule
GR ($S=2$) & $45.80$ & $27.79$ & $53.33$ & $42.31$ \\
\textbf{GR ($\bm{S=4}$)} & $47.68$ & $29.95$ & $54.20$ & $\mathbf{43.94}$ \\
GR ($S=8$) & $44.67$ & $26.65$ & $54.08$ & $41.80$\\
\bottomrule                         
\end{tabular}
}
\caption{Final performance of re-balancing variants on Qwen2-VL-7B-Instruct with $K=16$.}
\vspace{-5pt}
\label{tab:rebalancing variants}
\end{wraptable}

\paragraph{Variants of reshaping strategies.} We evaluate various re-balancing variants, with results presented in Table \ref{tab:rebalancing variants} (upper). First, experiments with TC reveal a trade-off in the choice of $L$: large $L$ inadequately alleviates data imbalance, whereas small $L$ fails to ensure sufficient diversity. 
We also explore \textbf{head clipping (HC)}, which removes fully-correct queries (i.e., $k_i=K$). Despite promising generalization capability, HC suffers from poor in-domain performance. 
Additionally, we test \textbf{repeat-based inverting (RI)}, a method that retains $K-k_i$ samples for each query $q_i\in \mathcal{D}_{\textnormal{filter}}$ and supplements the deficit through repetition. Compared to RP, RI yields marginally lower performance, emphasizing the importance of data diversity.

\paragraph{Variants of resampling strategies.}
Table \ref{tab:rebalancing variants} (lower) illustrates the performance of GR under different values of intermediate reasoning steps $S$. 
Too small $S$ ($S=2$) results in inadequate guidance effectiveness, while too large $S$ not only increases computational cost but also limits diversity, hindering further performance gains. This highlights the importance of selecting an appropriate value for $S$.

Furthermore, we compare our resampling strategies against brute-force resampling. As shown in Table \ref{tab:main}, AR and GR with $K=8$ achieve comparable or even superior performance to brute-force resampling (i.e., vanilla SI with $K=16$). 
% On Qwen2-VL-7B-Instruct model, AR and GR outperform brute-force resampling by $1.11$ and $0.97$ points on average, respectively. 
Regarding efficiency, brute-force resampling requires $K$ resampling, while AR reduces this to $K-k_i$ operations ($\approx 50\%$ cost reduction), and GR requires only $1$ resampling on minimal tail data. 
Focusing on data distribution, our strategies mitigate the Matthew effect in self-improvement, delivering superior performance with enhanced efficiency.

\subsection{Self-improvement as Efficient Sampling}
\begin{wraptable}{r}{10cm}
\vspace{-10pt}
\centering 
\resizebox{\linewidth}{!}{
\begin{tabular}{ccccc}
\toprule
\textbf{Method} & \textbf{MMPR} & \textbf{MathVerse} & \textbf{We-Math} & \textbf{Avg.} \\
\midrule  
batch sampling & $59.10$ & $33.12$ & $52.18$ & $48.13$ \\
\midrule
iterative sampling & $64.99$ & $31.47$ & $52.59$ & $49.68$ \\ 
+ TC & $66.62$ & $32.23$ & $\mathbf{53.10}$ & $50.65$ \\
+ RP & $\mathbf{69.89}$ & $\mathbf{33.88}$ & $50.63$ & $\mathbf{51.47}$ \\
\bottomrule                         
\end{tabular}
}
\caption{Performance comparison between batch sampling and iterative sampling on InternVL2.5-4B model.}
\label{tab:why_si}
\end{wraptable}
Driven by data distribution shifts during self-improvement, we hypothesize that inter-iteration sampling exhibits greater variance than intra-iteration sampling. To leverage this, we introduce \textbf{iterative sampling}, which combines $K=8$ samples over 5 iterations (40 samples total) to enhance data diversity. 
For comparison, we also implement \textbf{batch sampling}, which draws all $40$ samples at once. Results in Table \ref{tab:why_si} demonstrate that iterative sampling outperforms batch sampling, supporting our view of self-improvement as an efficient sampling method.
Furthermore, applying distribution-reshaping strategies to iterative sampling validates their effectiveness. Notably, RP outperforms both batch sampling and vanilla iterative sampling, with improvements of $3.34$ and $1.79$ points respectively.

\subsection{Self-correction Benefits Self-improvement}\label{sec:selfcorrection}

\begin{wrapfigure}{r}{10cm}
\centering
\vspace{-12pt}
\includegraphics[width=\linewidth]{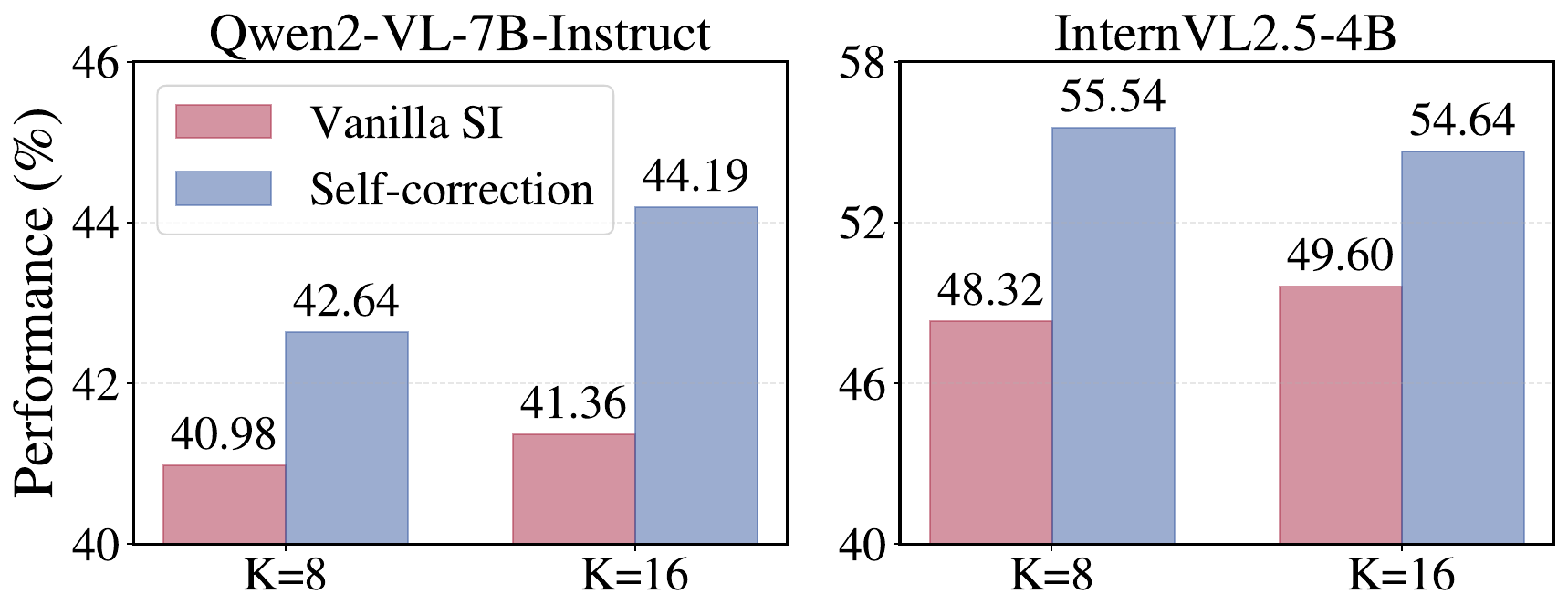}
\caption{Average performance comparison between vanilla and self-correction in visual self-improvement.}
\label{fig:selfcorrection}
\end{wrapfigure}

Self-correction has emerged as a promising learning paradigm \cite{DBLP:journals/corr/abs-2505-22651, Wu_2025_CVPR, DBLP:conf/naacl/ChengLXZZL25}, encouraging deeper reasoning and generating longer chains of thought. Therefore, we investigate its effectiveness for head-tail re-balancing in visual self-improvement (see Appendix \ref{appendix:F} for implementation details). In contrast to the resampling strategies discussed earlier, self-correction refines existing incorrect samples through $K-k_i$ operations per query, achieving efficiency comparable to AR and fully leveraging the potential of incorrect instances. Results in Figure \ref{fig:selfcorrection} demonstrate that self-correction effectively counteracts the Matthew effect, yielding substantial performance improvements.

\begin{wraptable}{r}{10cm}
\centering
\vspace{-10pt}
\resizebox{\linewidth}{!}{
\begin{tabular}{ccccc}
\toprule
\textbf{Method} & \textbf{MMPR} & \textbf{MathVerse} & \textbf{We-Math} & \textbf{Avg.} \\
\midrule
w/o CoT filtering & $44.04$ & $29.44$ & $\mathbf{55.11}$ & $42.87$ \\
w/ CoT filtering & $\mathbf{46.67}$ & $\mathbf{30.96}$ & $54.94$ & $\mathbf{44.19}$ \\ 
\bottomrule                         
\end{tabular}
}
\caption{Performance comparison between \textit{with} and \textit{without} CoT filtering on Qwen2-VL-7B-Instruct model.}
\vspace{-10pt}
\label{tab:self-correction}
\end{wraptable}

Notably, in addition to verifying final results, we also filter out data without CoT reasoning process. The results in the table \ref{tab:self-correction} indicate that CoT length filtering improved the quality of tail data, demonstrating superior overall performance.

\subsection{The Power of ``Seeing''}

\begin{wrapfigure}{r}{9cm}
\centering
\vspace{-12pt}
\includegraphics[width=\linewidth]{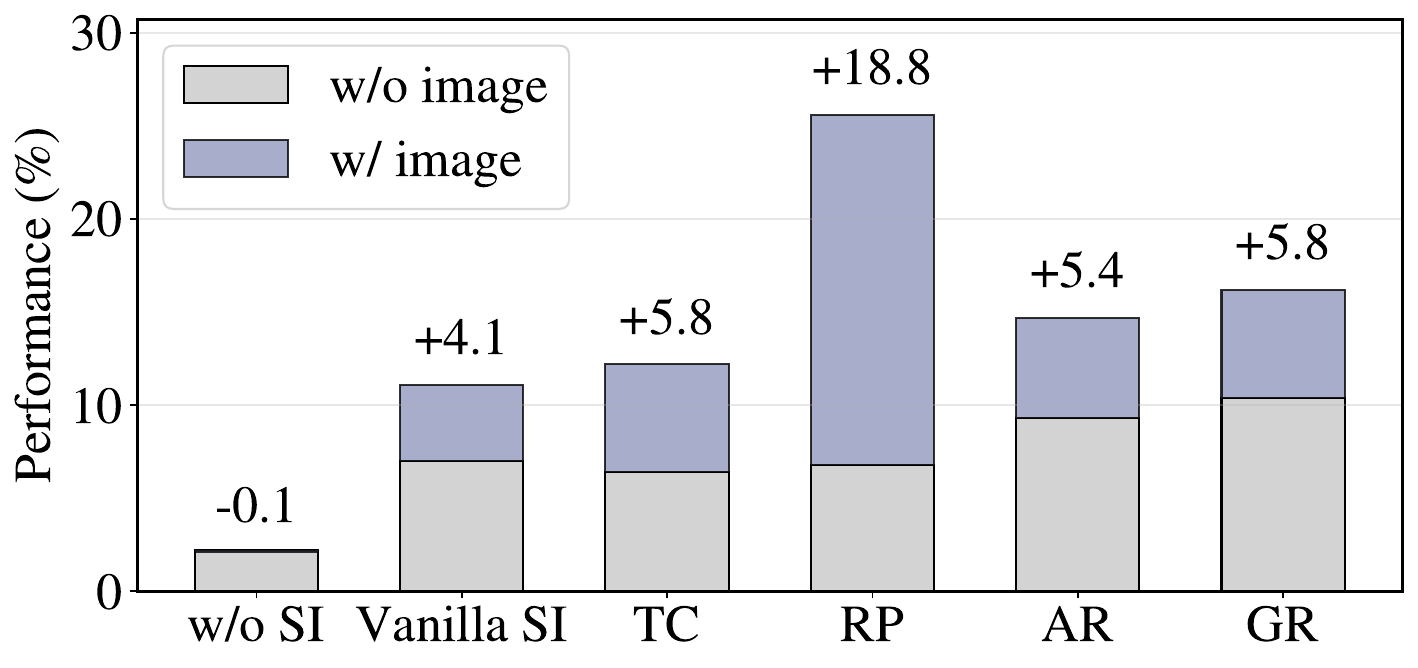}
\caption{Comparison of tail data performance \textit{with} and \textit{without} images on Qwen2-VL-7B-Instruct at $K=8$.}
\label{fig:with_wo_fig}
\end{wrapfigure}
To validate whether the final model truly enhances reasoning capabilities, we evaluate its performance on tail data under two settings: with and without image inputs. Figure \ref{fig:with_wo_fig} indicates that the base model exhibits negligible performance difference between these settings, indicating a poor capability to leverage visual information for solving challenging problems. In contrast, vanilla self-improvement leads to a substantial gain in real reasoning performance. Our re-balanced strategies further amplify this increment, with RP strategy demonstrating an $18.8$-point advantage when ``seeing'' images.

Additionally, for error type analysis and case study, please refer to Appendix \ref{appendix:B} and Appendix \ref{appendix:C}.

\section{Conclusion}
In this work, we identify a critical challenge behind performance bottlenecks in visual self-improvement: the ``Matthew effect'', where simple samples in the head progressively dominate successful trajectories, while difficult data in the tail becomes increasingly narrowing. To counteract it, we introduce four effective re-balanced strategies from distribution-reshaping and trajectory-resampling perspectives: threshold clipping, repeat-based padding, adaptive-weighted resampling, and guided resampling. Experimental results demonstrate that these strategies successfully reduce head dominance and increase tail proportion, thereby improving the performance ceilings. Future work will explore counteracting the Matthew effect on larger models and broader datasets, alongside developing more efficient re-balancing strategies.

\bibliography{main}

\clearpage
\newpage

\appendix
\section*{\centering \LARGE{Appendix}}

\section{Experimental Details}\label{appendix:A}
\subsection{Dataset Details}
We adopt MMPR \cite{DBLP:journals/corr/abs-2411-10442} as our primary dataset. From it, we randomly extract 7,980 mathematical reasoning samples \cite{DBLP:conf/acl/LuGJQHLZ20, DBLP:conf/coling/CaoX22, DBLP:conf/emnlp/SeoHFEM15} to construct MMPR-mini, with 7,183 for training and 797 for in-domain testing. 
Specifically, we select queries from Geometry3K \cite{DBLP:conf/acl/LuGJQHLZ20}, GeoQA+ \cite{DBLP:conf/coling/CaoX22}, and GEOS \cite{DBLP:conf/emnlp/SeoHFEM15} respectively, randomly sampling 10\% of data from each dataset to construct the test set. 
MMPR-mini primarily focuses on mathematical visual reasoning problems, including multiple-choice questions, open-ended questions, and other formats, containing only queries with their corresponding ground truth without CoT reasoning trajectories.

Additionally, for out-of-domain (OOD) evaluation, we select two widely-adopted mathematical visual reasoning datasets: MathVerse \cite{DBLP:conf/eccv/ZhangJZLGQZLCQGL24} and We-Math \cite{DBLP:journals/corr/abs-2407-01284}, comprising 788 and 1,740 queries respectively. All datasets we utilized are open-source.

\subsection{Difficulty Level Categorization}
Using Qwen2-VL-7B-Instruct, we perform 64-shot sampling on each query, and categorize the queries into 5 difficulty levels based on their pass@64 performance. This classification prioritizes a balanced data distribution across all levels, with data difficulty ascending from level 1 to level 5.

\subsection{Prompt Details}

\begin{figure}[h!]
\centering
\includegraphics[width=0.55\columnwidth]{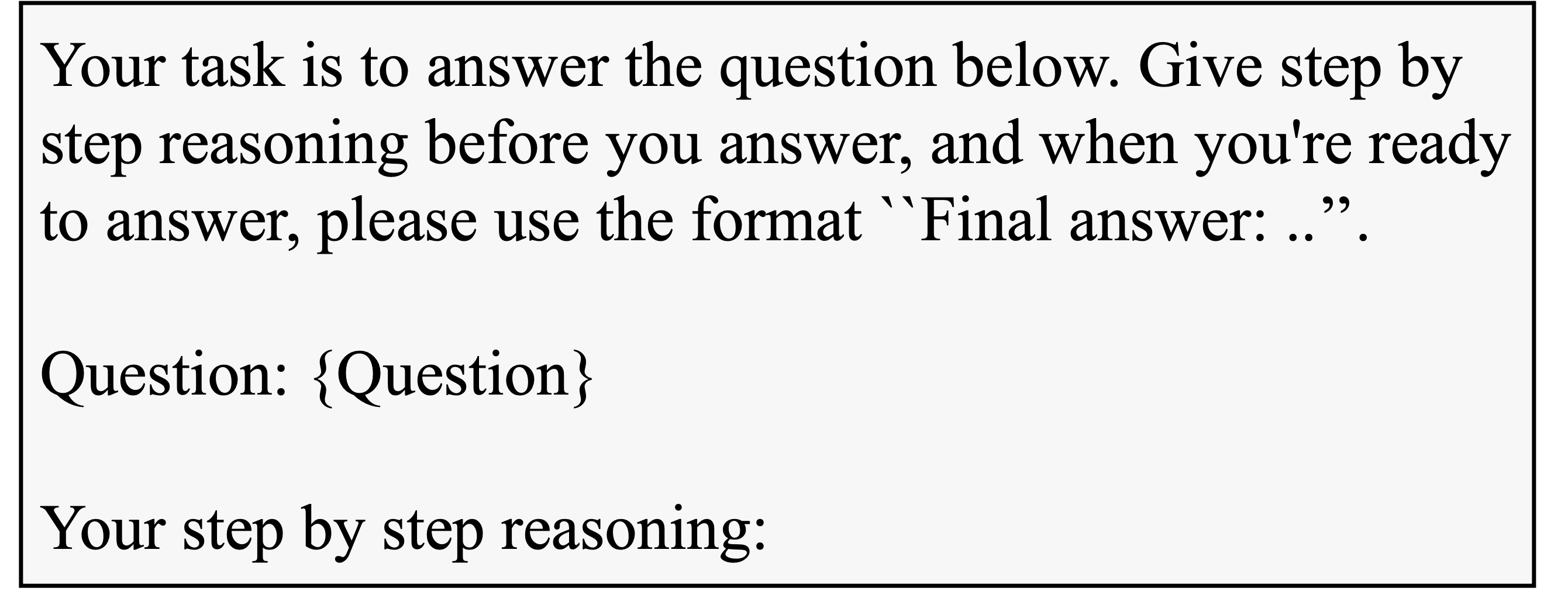}
\caption{Prompt for training, sampling and testing.}
\label{fig:prompt}
\end{figure}

In this work, we use the unified prompt template (see Figure \ref{fig:prompt}) in the phase of training, sampling and testing across varying datasets.

\section{Error Type Analysis}\label{appendix:B}

First, we analyze the incorrect data and categorize it into the following error types:

\begin{itemize}[leftmargin=*]
\item \textbf{No Reasoning Process (NP)}: Models provide answers directly without demonstrating any step-by-step reasoning or explanation process.
\item \textbf{Comprehension Error (CE)}: Models exhibit misunderstanding of either the question content or the visual information presented in the image.
\item \textbf{Knowledge Error (KE)}: Models employ incorrect formulas, theorems, or other factual information.
\item \textbf{Logic Error (LE)}: Models generate flawed reasoning steps, perform incorrect calculations, or establish faulty cause-and-effect relationships.
\item \textbf{Format Error (FE)}: Models produce responses whose format does not meet the specified requirements.
\end{itemize}

\begin{figure}[h!]
\centering
\includegraphics[width=0.55\columnwidth]{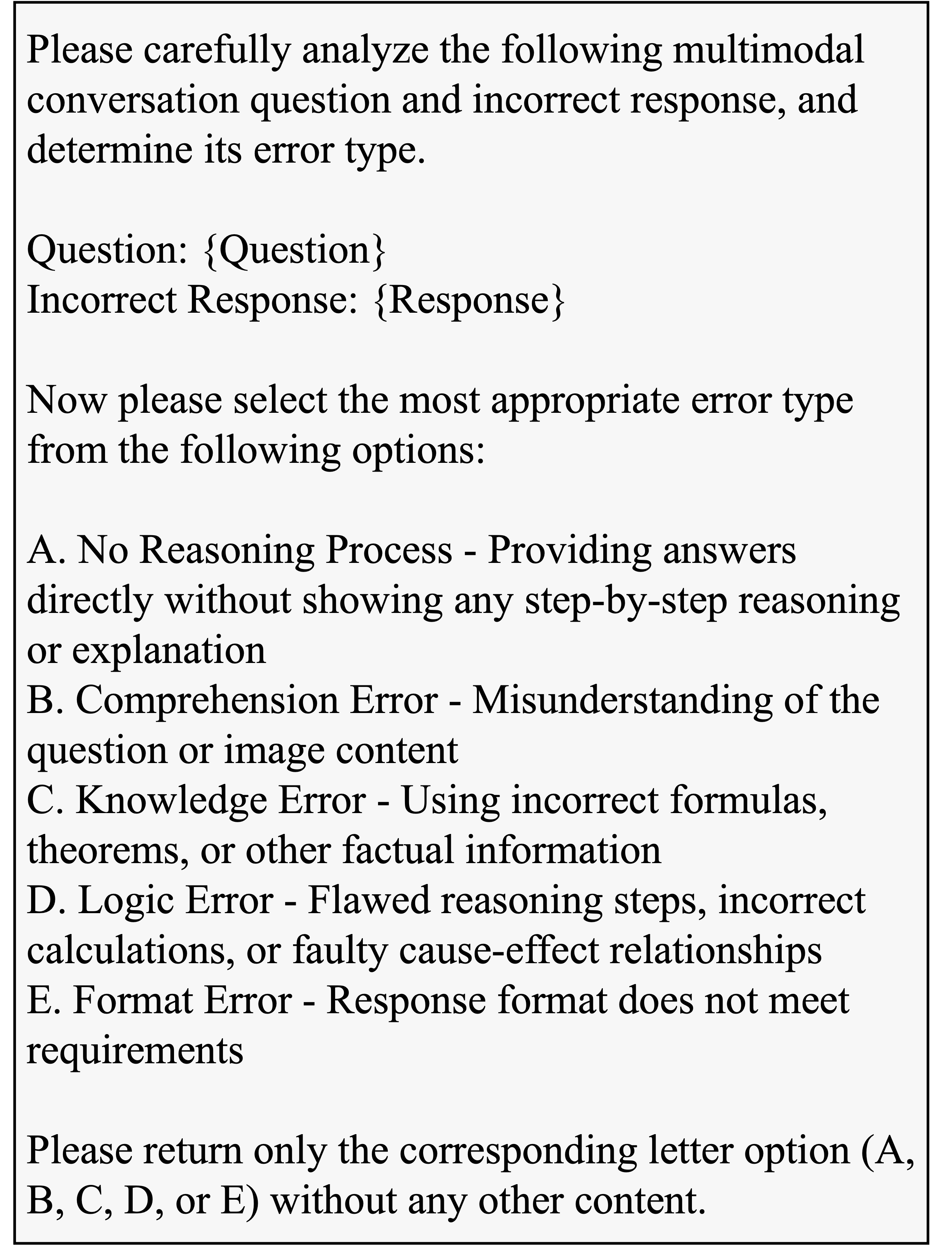}
\caption{Prompt for determining error type.}
\label{fig:prompt_type}
\end{figure}

Then we utilize GPT-4.1 \cite{DBLP:journals/corr/abs-2303-08774} for extensive error type classification, setting temperature=0.1 and permitting up to 3 outputs. The prompt template is illustrated in Figure \ref{fig:prompt_type}, with classification outcomes detailed in Table \ref{tab:error}. 

\begin{table}[ht]
\centering 
\resizebox{0.5\linewidth}{!}{
\begin{tabular}{ccccccc}
\toprule
\textbf{Model} & \textbf{Method} & \textbf{NP} & \textbf{CE} & \textbf{KE} & \textbf{LE} & \textbf{FE} \\ 
\midrule
\multirow{6}{*}{Qwen2-VL} & w/o SI & 4 & 286 & 24 & 286 & 2 \\
 & Vanilla SI & 36 & 189 & 21 & 216 & 1 \\ 
 & TC & 182 & 137 & 16 & 91 & 0 \\  
 & RP & 34 & 152 & 10 & 222 & 0 \\ 
 & AR & 77 & 185 & 13 & 170 & 2 \\ 
 & GR & 153 & 128 & 8 & 124 & 2 \\
\midrule
\multirow{6}{*}{InternVL2.5} & w/o SI & 0 & 227 & 13 & 176 & 0 \\
 & Vanilla SI & 1 & 132 & 12 & 135 & 2 \\ 
 & TC & 0 & 150 & 12 & 123 & 0 \\  
 & RP & 14 & 123 & 9 & 116 & 0 \\ 
 & AR & 7 & 137 & 12 & 113 & 0 \\ 
 & GR & 0 & 126 & 14 & 117 & 1 \\ 
\bottomrule                         
\end{tabular}
}
\caption{Error type analysis. NP, CE, KE, LE and FE denote No Reasoning Process, Comprehension Error, Knowledge Error, Logic Error, and Format Error, respectively. SI, TC, RP, AR, and GR denote self-improvement, threshold clipping, repeat-based padding, adaptive-weighted resampling, and guided resampling, respectively.}
\label{tab:error}
\end{table}

The results reveal several key findings: (1) The overall distribution of error types varies significantly across different models. For instance, with the Qwen2-VL-7B-Instruct model, the number of samples directly providing final answers increases substantially, particularly under the TC strategy, which reduces the quantity of data and further amplifies the proportion of direct-answer samples, making ``No Reasoning Process'' the most frequent error type in TC's incorrect data. In contrast, the InternVL2.5-4B model exhibits few "No Reasoning Process" errors. (2) Even without self-improvement, format errors occur with extremely low frequency, indicating that the self-improvement process truly enhances models' reasoning capabilities rather than simple instruction-following abilities for format. (3) Self-improvement achieves significant improvements mainly in comprehension errors and logic errors, substantially reducing the occurrence of both error types. (4) Compared to vanilla self-improvement, our re-balanced strategies enhance models' capabilities in question and image comprehension as well as reasoning logic.

\begin{figure}[h!]
\centering
\includegraphics[width=0.55\columnwidth]{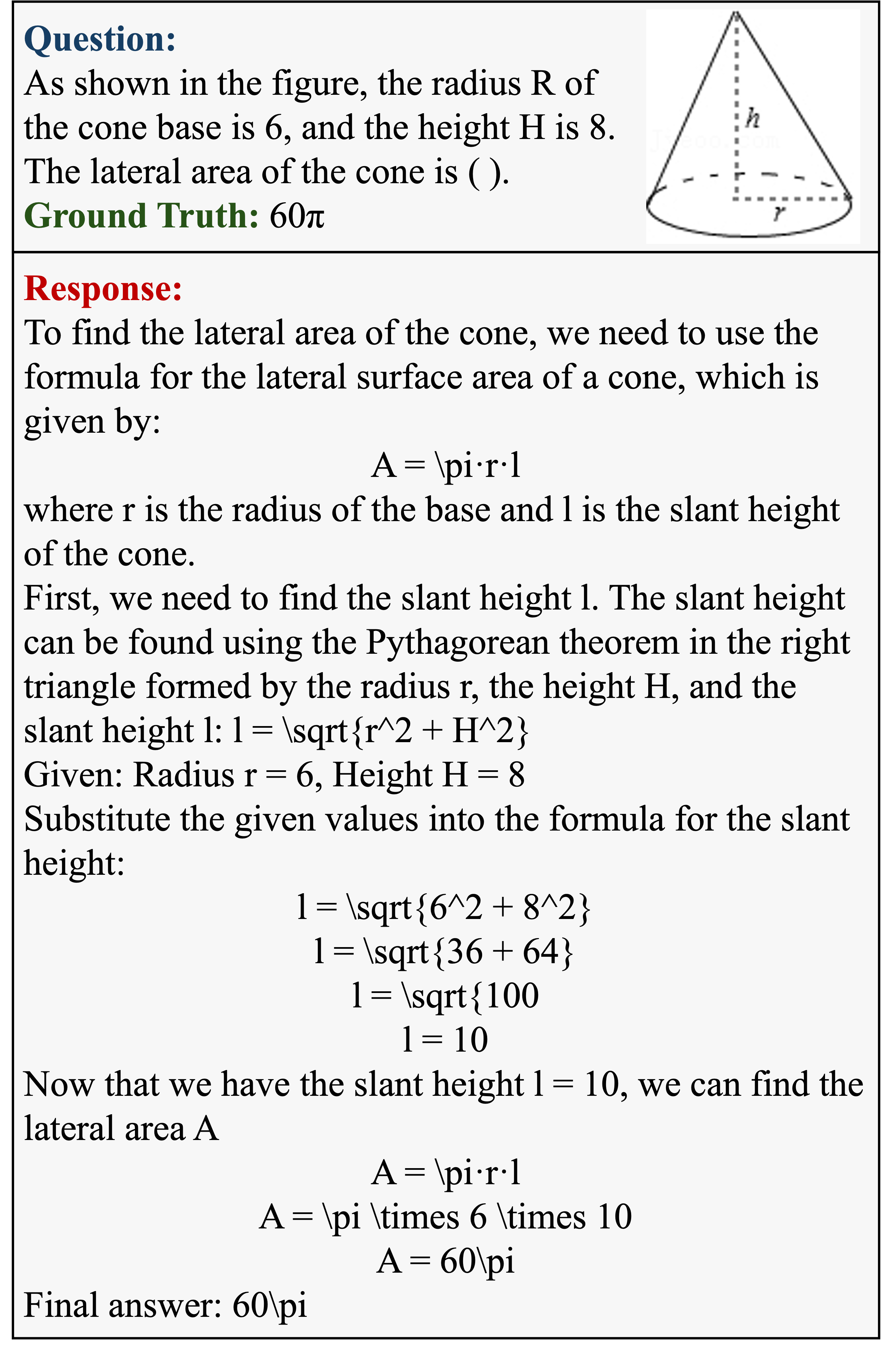}
\caption{A case misclassified as incorrect due to exact-match rules, where \textbackslash pi and $\pi$ appear in different formats.}
\label{fig:case_right}
\end{figure}

In rare instances, the model consistently returns ``T''; these cases are omitted from the table. Manual examination of these cases reveals that the responses are accurate but were misclassified as incorrect due to our exact-match rules, with an illustrative example provided in Figure \ref{fig:case_right}. 

\section{Case Study}\label{appendix:C}
\begin{figure}[h!]
\centering
\includegraphics[width=0.6\columnwidth]{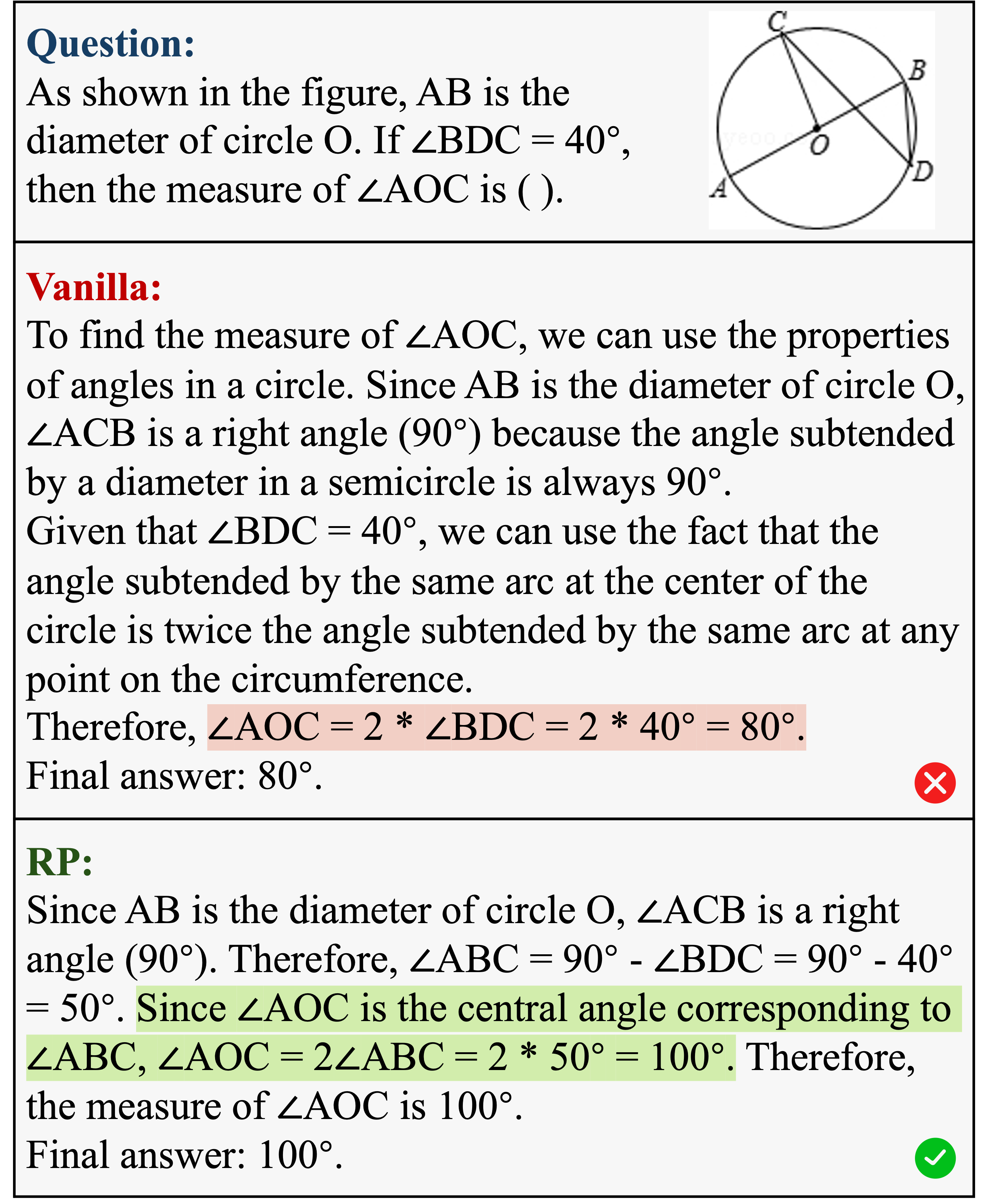}
\caption{An example of repeat-based padding with Qwen2-VL-7B-Instruct model on MMPR-mini test set. 
Compared to vanilla self-improvement, repeat-based padding successfully finds the relationship between $\angle$AOC and $\angle$ABC, reaching the right answer.
}
\label{fig:case_rp}
\end{figure}

\begin{figure}[h!]
\centering
\includegraphics[width=0.55\columnwidth]{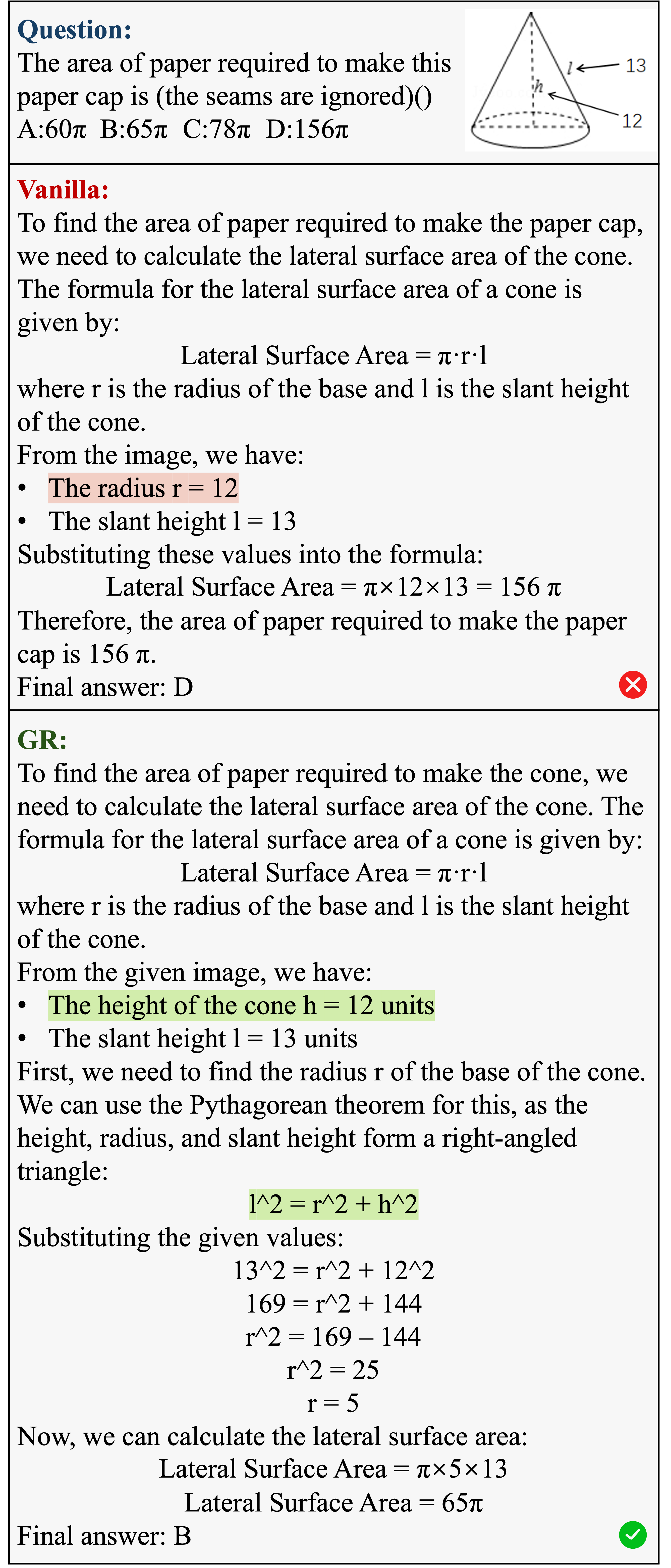}
\caption{An example of guided resampling with InternVL2.5-4B model on MathVerse test set. In this case, vanilla self-improvement demonstrates the confusion of height $h$ with radius $r$, while guided resampling addresses this problem successfully.
}
\label{fig:case_gr}
\end{figure}

Our main results show that repeat-based padding and guided resampling achieve superior performance among our proposed strategies. Therefore, we showcase examples of these two strategies in Figure \ref{fig:case_rp} and \ref{fig:case_gr} respectively. 
Both cases reveal that vanilla self-improvement exhibits limited capabilities in visual comprehension and geometric element understanding. 
For instance, in Figure \ref{fig:case_rp}, vanilla self-improvement incorrectly determines the relationship between $\angle$AOC and $\angle$BDC, while in Figure \ref{fig:case_gr} demonstrates the confusion of height $h$ with radius $r$, resulting in final errors. Both repeat-based padding and guided resampling successfully address these problems.

\section{Matthew Effect Mitigation}\label{appendix:D}
\begin{figure}[h!]
\centering
\includegraphics[width=0.55\columnwidth]{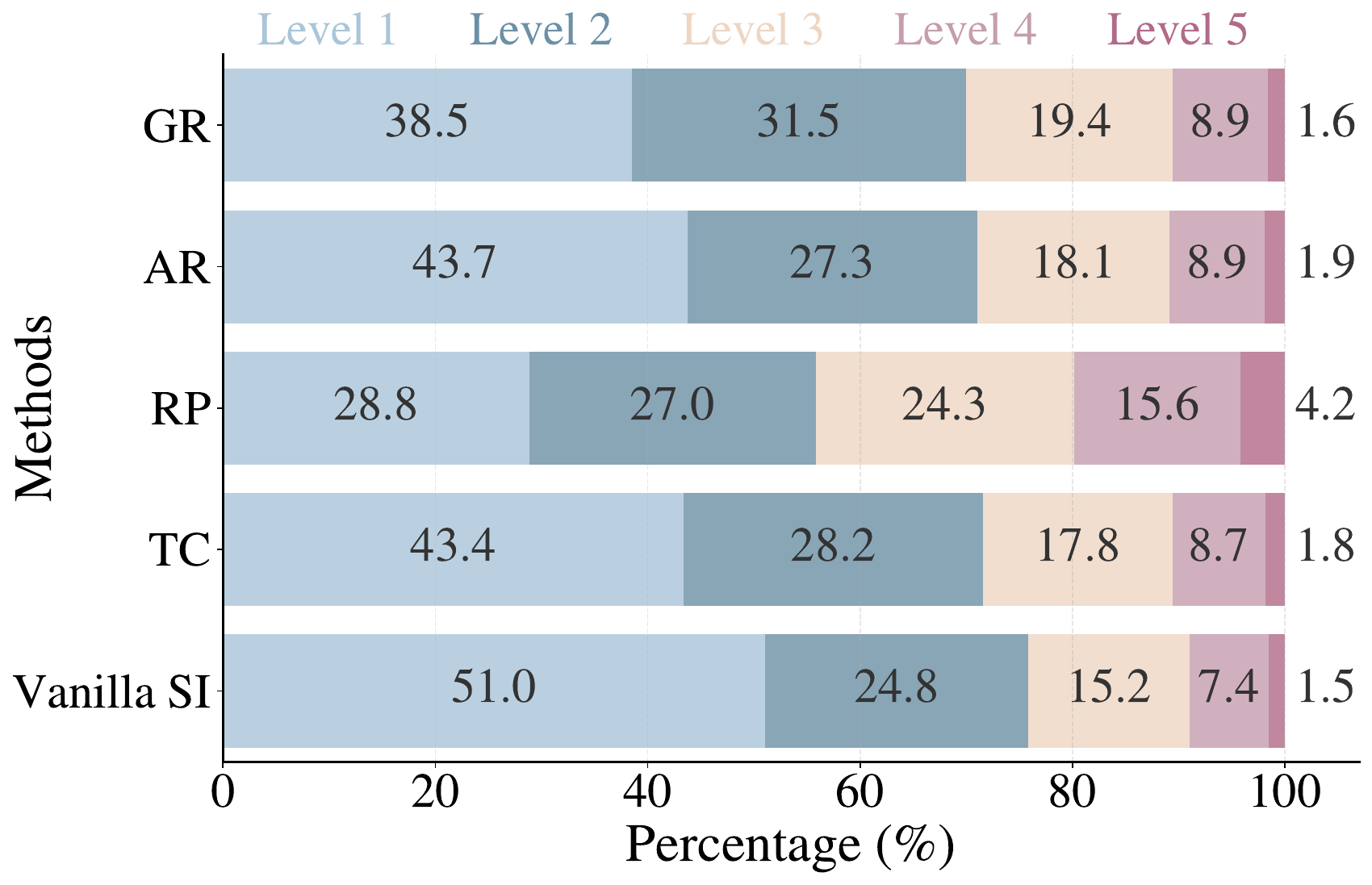}
\caption{Data distribution of difficulty levels in successful trajectories under different strategies with Qwen2-VL-7B-Instruct at $K=8$.}
\label{fig:strategies_qwen_k8}
\end{figure}

\begin{figure}[h!]
\centering
\includegraphics[width=0.55\columnwidth]{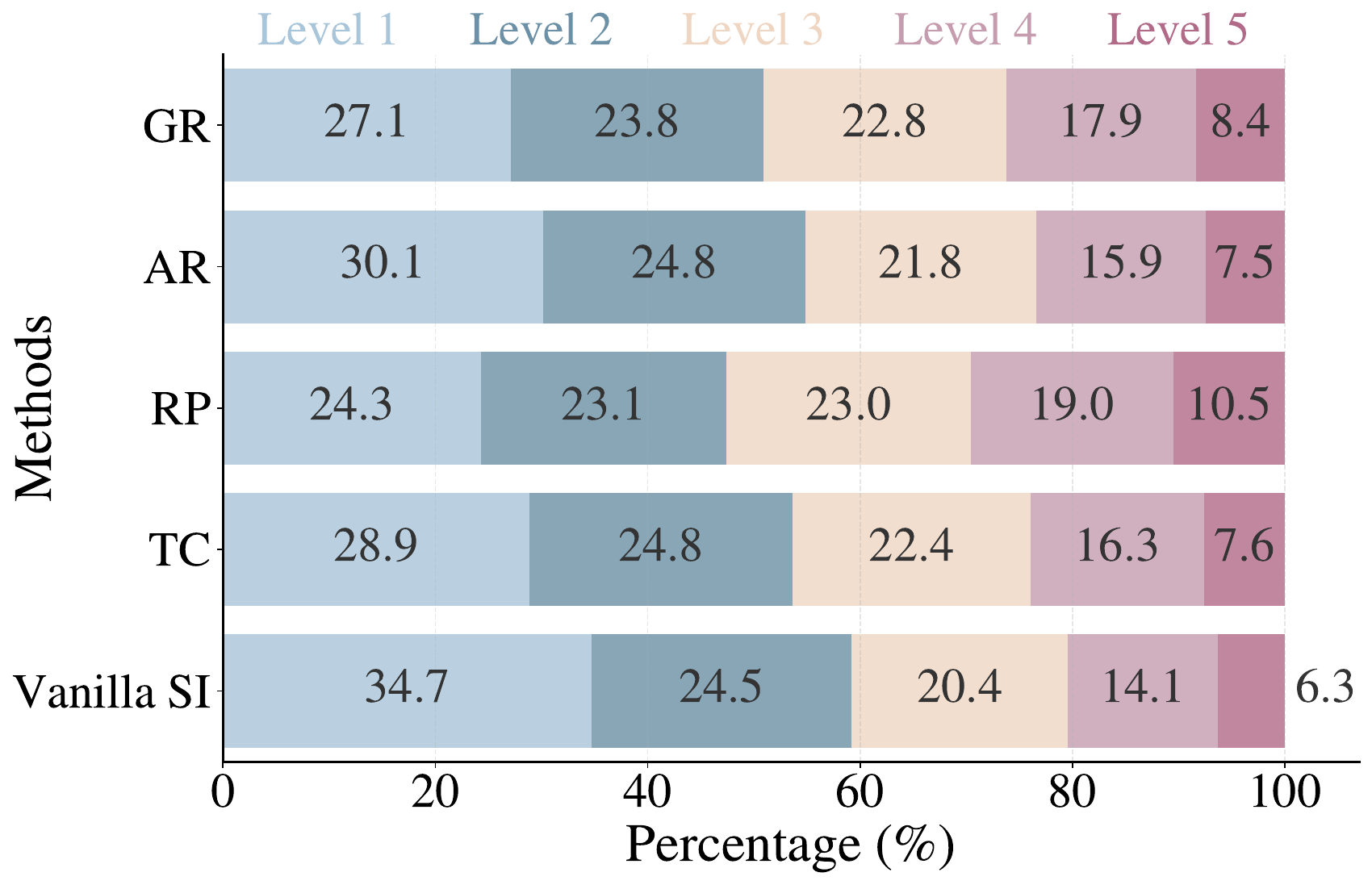}
\caption{Data distribution of difficulty levels in successful trajectories under different strategies with InternVL2.4-4B at $K=8$.}
\label{fig:strategies_internvl_k8}
\end{figure}

\begin{figure}[h!]
\centering
\includegraphics[width=0.55\columnwidth]{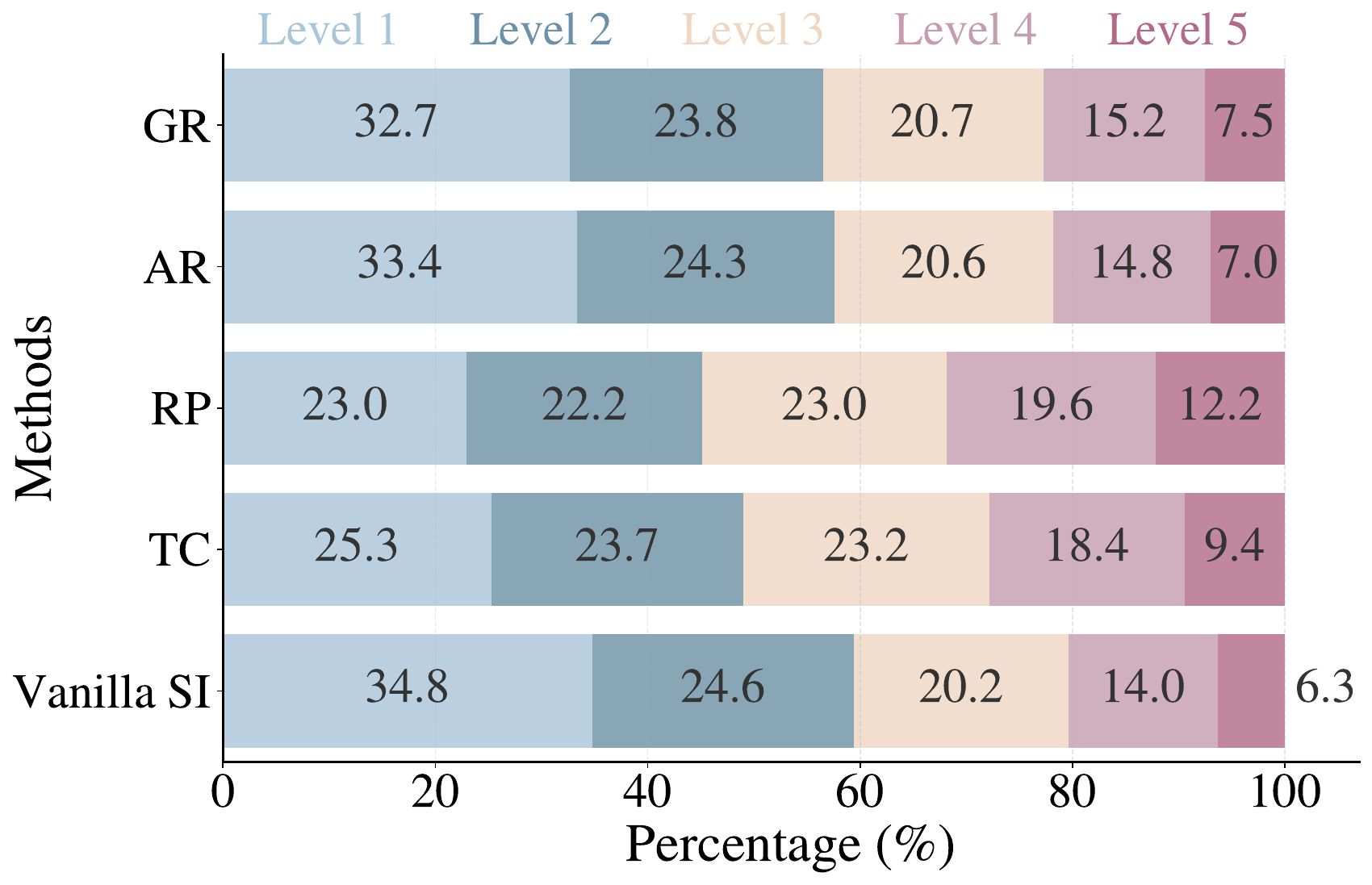}
\caption{Data distribution of difficulty levels in successful trajectories under different strategies with InternVL2.4-4B at $K=16$.}
\label{fig:strategies_internvl_k16}
\end{figure}

In Section \ref{sec:experiments}, we analyzed the effectiveness of our proposed rebalanced strategies in mitigating the Matthew effect and their performance on visual reasoning tasks using Qwen2-VL-7B-Instruct at $K=16$. Here, we present three additional configurations–Qwen2-VL-7B-Instruct at $K=8$, InternVL2.5-4B at $K=8$, and InternVL2.5-4B at $K=16$–in Figure \ref{fig:strategies_qwen_k8}, Figure \ref{fig:strategies_internvl_k8} and Figure \ref{fig:strategies_internvl_k16}.

Results demonstrate that our proposed re-balancing strategies effectively mitigate the Matthew effect in self-improvement across different experimental settings. Overall, the repeat-based padding (RP) method exhibits the best mitigation performance. Among the two trajectory-resampling strategies, guided resampling (GR) generally outperforms adaptive-weighted resampling (AR), which aligns with the performance patterns of these strategies on the test sets.

\section{Batch Sampling and Iterative Sampling}\label{appendix:E}

In Section \ref{sec:discussion}, we used InternVL2.5-4B at $K=8$ as an example to discuss that self-improvement can be viewed as an efficient sampling approach. Additionally, we conducted similar experiments on Qwen2-VL-7B-Instruct at $K=16$ and observed similar phenomena. Specifically, we combine $K=16$ samples across 5 iterations (totaling 80 samples) as iterative sampling, while batch sampling involves directly sampling 80 times from the base model. As shown in Table \ref{tab:why_si_appendix}, iterative sampling delivers improvements of $5.65$ points on the in-domain test set and $3.17$ points on average. Nevertheless, our distribution-reshaping strategies do not consistently perform well under the iterative sampling perspective. This might be because for $K=16$, TC with $L=4$ significantly reduces the data quantity, while RP leads to excessive repetition, both of which may hinder advantages in efficient sampling scenarios.

\begin{table}[h!]
\centering 
\resizebox{0.65\linewidth}{!}{
\begin{tabular}{ccccc}
\toprule
\textbf{Method} & \textbf{MMPR} & \textbf{MathVerse} & \textbf{We-Math} & \textbf{Avg.} \\
\midrule  
batch sampling & $38.52$ & $26.52$ & $49.37$ & $38.14$ \\
\midrule
iterative sampling & $\mathbf{44.17}$ & $\mathbf{28.68}$ & $51.15$ & $\mathbf{41.33}$ \\ 
+ TC & $41.78$ & $27.41$ & $\mathbf{52.24}$ & $40.48$ \\
+ RP & $42.03$ & $26.52$ & $48.45$ & $39.00$ \\
\bottomrule                         
\end{tabular}
}
\caption{Performance comparison between batch sampling and iterative sampling on Qwen2-VL-7B-Instruct model.}
\label{tab:why_si_appendix}
\end{table}

\section{Applying Self-correction to Self-improvement}\label{appendix:F}
\begin{figure}[h!]
\centering
\includegraphics[width=0.65\columnwidth]{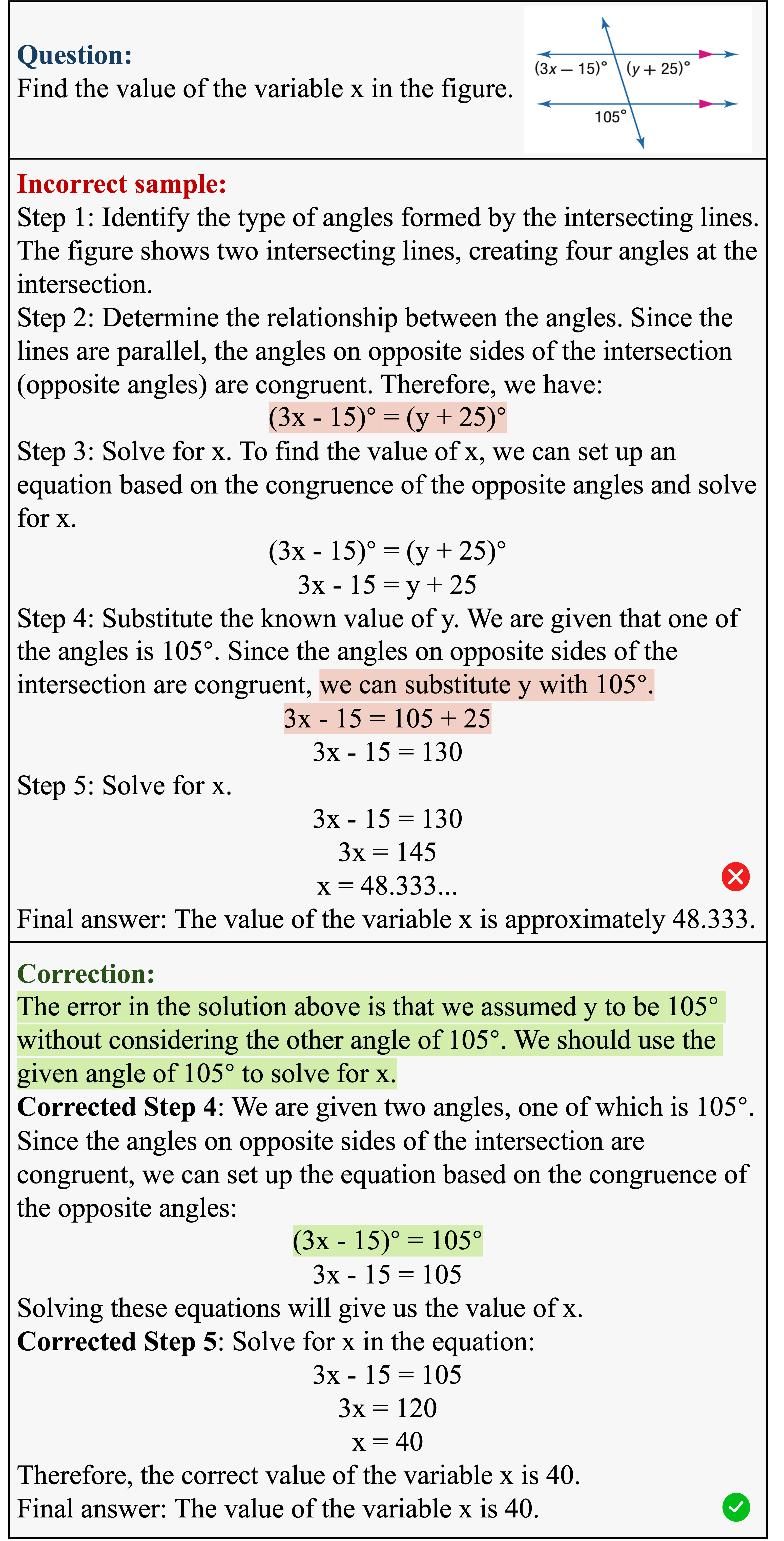}
\caption{An example of self-correction capabilities of Qwen2-VL-7B-Instruct model.}
\label{fig:case_sc}
\end{figure}

Given self-correction capabilities shown in Figure \ref{fig:case_sc}, we explore applying self-correction to self-improvement for head-tail re-balancing, incorporating the refined outputs into subsequent training iterations. 

\begin{figure}[h!]
\centering
\includegraphics[width=0.6\linewidth]{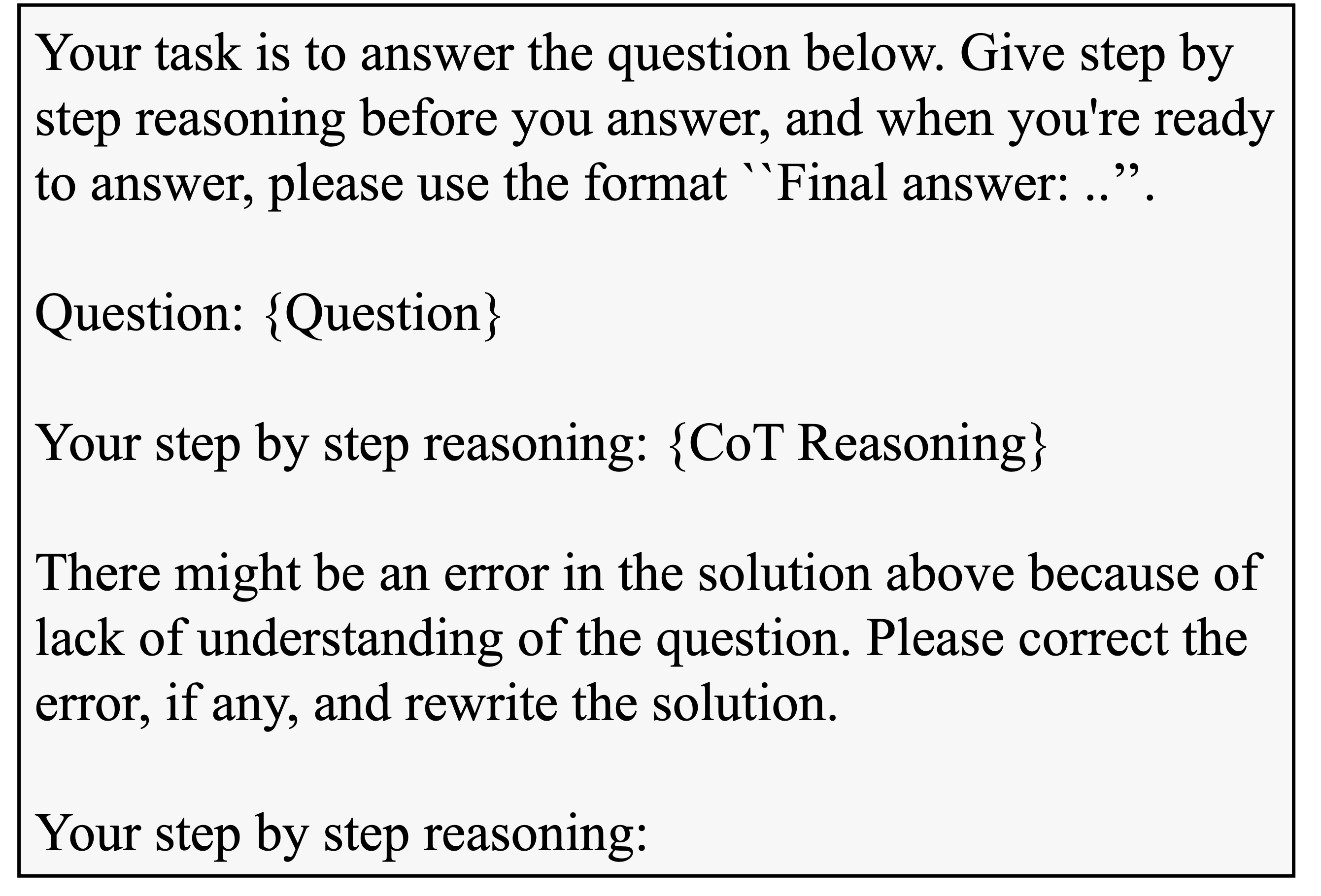}
\caption{Prompt for self-correction.}
\label{fig:prompt_selfcorrection}
\end{figure}

Formally, we first define the set of incorrect samples as: $$\mathcal{D}_{\textnormal{discard}}^{(t)}=\{(q_i, \hat{r}_{i,k}^{(t)})\in \mathcal{D}_{\textnormal{sample}}^{(t)}\ |\ rf(q_i, a_i, \hat{a}_{i,k}^{(t)})=0\}.$$
For $\hat{r}_i\in\mathcal{D}_{\textnormal{discard}}^{(t)}$, we design the self-correction prompt $p$ to resample $\tilde{r}_{i}\sim\mathcal{M}_{t-1}(\cdot|q_i,\hat{r}_i,p)$ and form the resample dataset:
$$\mathcal{D}_{\textnormal{self-correction}}^{(t)}=\{(q_i, \hat{r}_{i, k}^{(t)}, p, \tilde{r}_{i, k}^{(t)}), (q_i, \tilde{r}_{i,k}^{(t)})\ |\ k_i<K\},$$
where we filter out samples with failed corrections and insufficient CoT reasoning length. 
Finally, we obtain the training dataset as following:
$$\mathcal{D}_{\textnormal{train-BP}}^{(t)}=\mathcal{D}_{\textnormal{train}}^{(t)}\cup\mathcal{D}_{\textnormal{self-correction}}^{(t)}\ .$$
The prompt $p$ we used \cite{DBLP:conf/iclr/KumarZASCSBIBRZ25} is presented in Figure \ref{fig:prompt_selfcorrection}.

\begin{figure*}[h!]
\centering
\includegraphics[width=0.9\textwidth]{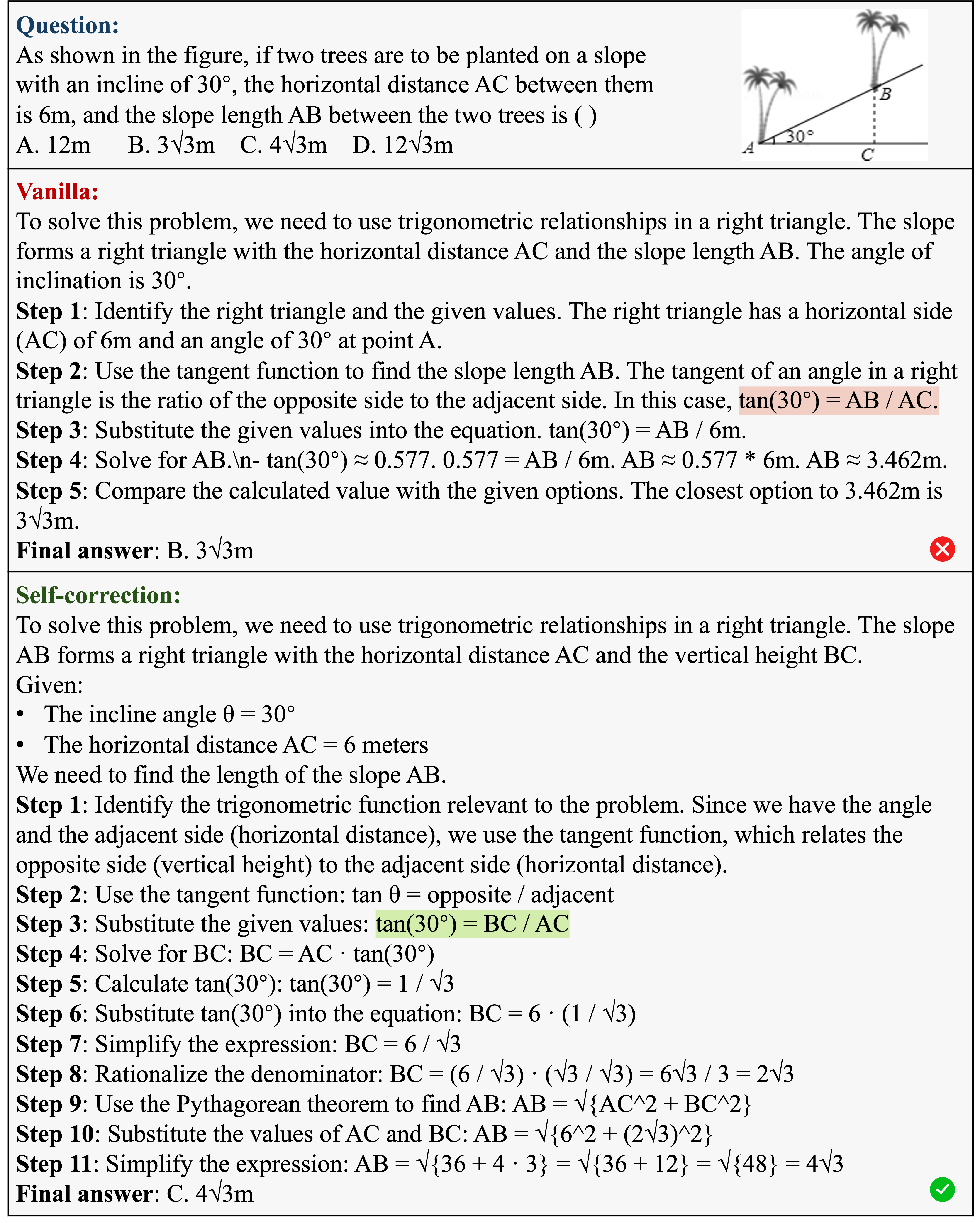}
\caption{An example of self-correction with InternVL2.5-4B model on MMPR-mini test set. In this case, the model of vanilla self-improvement made identification errors during tangent calculations, indicating deficient visual comprehension capabilities. In contrast, the self-correction method successfully addresses this problem and performs detailed computations to reach the correct answer.}
\label{fig:case_sc_right}
\end{figure*}

Results in Section \ref{sec:selfcorrection} demonstrate that applying self-correction to self-improvement yields substantial performance gains. 
We provide an illustrative example in Figure \ref{fig:case_sc_right}, where iterative self-correction enables the model to engage in deeper reasoning through generating longer thinking chains, ultimately correcting errors in vanilla self-improvement.

\end{document}